\documentclass[journal]{IEEEtran}

\usepackage{amsfonts}	% Additional math fonts
\usepackage{amsmath}	% Math symbols
\usepackage{amssymb}    % Extended math symbols
\usepackage{xfrac}    % nice fractions in text
\usepackage{siunitx}
\usepackage{mathtools}
\usepackage{pifont}   % check mark
\usepackage{dsfont}   % indicator function

\usepackage{cuted} % supplement

\usepackage{booktabs}
\usepackage{makecell}  % Line breaks in table cells
\usepackage[flushleft]{threeparttable}  % For notes below a table
\usepackage{multirow}
\usepackage{colortbl}

\usepackage{xspace}    % for correct spacing after defined commands without arguments, e.g., \net
\usepackage[table]{xcolor}
\usepackage[inline]{enumitem} % To change label for enumerates
\usepackage{textcomp}  % prevent warnings of gensymb package
\usepackage{gensymb}   % \degree
\usepackage{graphicx} % insert PDF as figure
\usepackage[caption=false,font=footnotesize]{subfig}
\usepackage{microtype}
\usepackage{cite}
\usepackage{flushend}
\usepackage{placeins} % \FloatBarrier
\usepackage{url}

\newcommand{\cmark}{\ding{51}}%
\newcommand{\xmark}{\ding{55}}%

\usepackage{stfloats} 

\makeatletter
\let\NAT@parse\undefined
\makeatother

\usepackage[pdfencoding=auto, hidelinks]{hyperref} % Usually, this is the last package to be imported
\usepackage[hang,flushmargin]{footmisc}

\usepackage[capitalize]{cleveref}
\crefname{section}{Sec.}{Secs.}
\Crefname{section}{Section}{Sections}
\Crefname{table}{Table}{Tables}
\crefname{table}{Tab.}{Tabs.}

\newcommand{\net}{\mbox{DualViewDistill}\xspace}

\definecolor{Gray}{gray}{0.9}

\begin{document}

\title{Bridging Perspectives: Foundation Model Guided BEV Maps for 3D Object Detection and Tracking}

\author{
Markus Käppeler$^{1}$,
Özgün Çiçek$^{2}$,
Daniele Cattaneo$^{1}$,
Claudius Gläser$^{2}$,
Yakov Miron$^{2}$,
and Abhinav Valada$^{1}$% <-this % stops a space
%\thanks{$^{1}$ Markus Käppeler, Daniele Cattaneo, and Abhinav Valada are with the Department of Computer Science, University of Freiburg, Germany.}%
%\thanks{$^{2}$ Özgün Çiçek, Yakov Miron, and Claudius Gläser are with Bosch Research, Robert Bosch GmbH, Renningen, Germany.}
\thanks{$^{1}$ Department of Computer Science, University of Freiburg, Germany.}%
\thanks{$^{2}$ Bosch Research, Robert Bosch GmbH, Renningen, Germany.}%
\thanks{This research was funded by Bosch Research as part of a collaboration between Bosch Research and the University of Freiburg on AI-based automated driving.}%
}

% Paper headers

\maketitle

\bstctlcite{BSTcontrol}

%%%%%%%%%%%%%%%%%%%%%%%%%%%%%%%%%%%%%%%%%%%%%%%%%%%%%%%%%%%%%%%%%%%%%%%%%%%%%%%%

\begin{abstract}
    Camera-based 3D object detection and tracking are essential for perception in autonomous driving. Current state-of-the-art approaches often rely exclusively on either perspective-view (PV) or bird’s-eye-view (BEV) features, limiting their ability to leverage both fine-grained object details and spatially structured scene representations. In this work, we propose \net, a hybrid detection and tracking framework that incorporates both PV and BEV camera image features to leverage their complementary strengths. Our approach introduces BEV maps guided by foundation models, leveraging descriptive DINOv2 features that are distilled into BEV representations through a novel distillation process. By integrating PV features with BEV maps enriched with semantic and geometric features from DINOv2, our model leverages this hybrid representation via deformable aggregation to enhance 3D object detection and tracking. Extensive experiments on the nuScenes and Argoverse~2 benchmarks demonstrate that \net achieves state-of-the-art performance. The results showcase the potential of foundation model BEV maps to enable more reliable perception for autonomous driving. We make the code and pre-trained models available at \mbox{\url{https://dualviewdistill.cs.uni-freiburg.de}}.

\end{abstract}

%%%%%%%%%%%%%%%%%%%%%%%%%%%%%%%%%%%%%%%%%%%%%%%%%%%%%%%%%%%%%%%%%%%%%%%%%%%%%%%%

%\begin{IEEEkeywords}
%Computer Vision for Transportation; Object Detection, Segmentation and Categorization; Deep Learning in Robotics and Automation; 3D Object Detection and Tracking
%3D Object Detection, Multi-Object Tracking, Foundation Models
%\end{IEEEkeywords}

%%%%%%%%%%%%%%%%%%%%%%%%%%%%%%%%%%%%%%%%%%%%%%%%%%%%%%%%%%%%%%%%%%%%%%%%%%%%%%%%

\section{Introduction}

\IEEEPARstart{C}{amera-based} 3D object detection and multi-object tracking are fundamental tasks in the autonomous driving pipeline, enabling reliable perception for downstream prediction and planning. Recent state-of-the-art methods can be categorized into two main approaches: perspective-view (PV)-based and bird’s-eye-view (BEV)-based detection and tracking. PV-based methods~\cite{lin2023sparse4d, wang2023exploring, jiang2024far3d, buchner20223d} rely on camera image features directly, while BEV-based approaches~\cite{li2024bevnext, han2024exploring, Park2022TimeWT, yang2023bevformer, huang2022bevpoolv2} transform camera features into the top-down BEV metric representation. This top-down BEV representation enhances spatial reasoning by mapping multiple camera images into a unified space, but potentially discarding fine-grained details during the transformation.

In parallel, vision foundation models such as DINOv2~\cite{oquab2023dinov2}, CLIP~\cite{radford2021learning}, and Stable Diffusion~\cite{rombach2022high} have demonstrated remarkable generalization capabilities across diverse vision tasks. Recent works~\cite{sirko2024occfeat,wang2024distillnerf} leverage these foundation models to distill their general-purpose features into 3D space, enabling label-efficient BEV and 3D occupancy semantic segmentation. However, such distilled descriptive scene representations have yet to be leveraged for supervised 3D object detection and tracking. Additionally, LiDAR-based CNN detectors~\cite{yang2018hdnet,fang2021mapfusion,huang2023menet} incorporate scene context via rasterized BEV HD maps to enhance object detection. However, these methods rely on manually labeled HD maps, which are difficult to obtain, and have not been effectively adopted for the camera-only setting.

\begin{figure}[t]
    \centering
    \includegraphics[width=\linewidth]{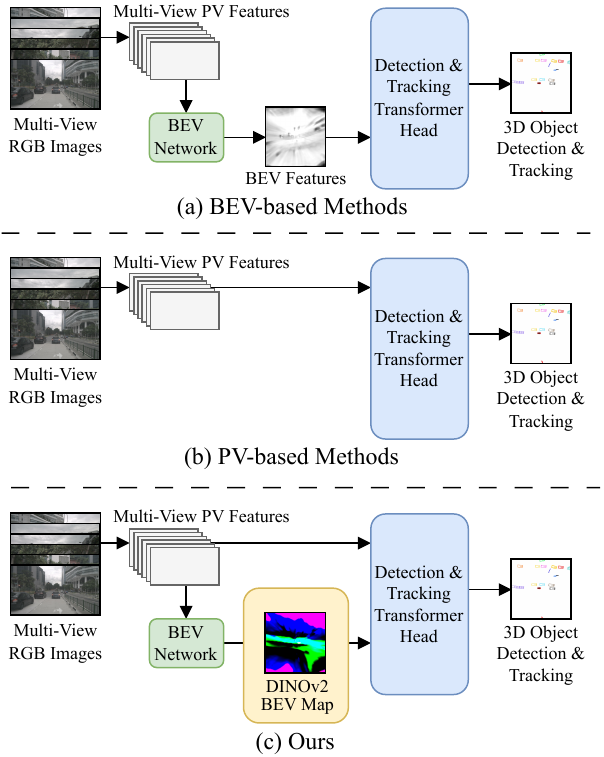}
    \caption{\net (c) integrates both perspective view (PV) features and latent bird’s-eye-view (BEV) features guided by DINOv2 features, to improve 3D object detection and tracking, while previous methods either rely only on BEV (a) or PV (b) features.}
    \label{fig:cover}
\end{figure}

Despite these advancements, current state-of-the-art methods exhibit several limitations, as illustrated in Fig.~\ref{fig:cover}. First, existing approaches exclusively rely on either BEV or PV features. BEV-based models benefit from a structured metric representation that encodes and stores static surroundings but lack direct access to multi-scale PV features, making them less effective at capturing object-centric details. Conversely, PV-based methods operate in perspective space, limiting their capacity to store spatially unified scene information effectively. To address these limitations, we propose leveraging both PV and BEV camera image features for 3D object detection and tracking, combining the strengths of both representations into a hybrid model.

Second, existing map-based detectors~\cite{yang2018hdnet, fang2021mapfusion, huang2023menet} utilize a fixed set of semantic BEV map classes to incorporate scene context and prior map information. However, foundation models offer rich feature representations with more general semantic and geometric information, capturing a broader range of semantic categories in their latent feature space.
To exploit this potential, we propose distilling DINOv2 features into BEV representations. Specifically, we first extract DINOv2 features for each available camera image and project these features onto the LiDAR point cloud of the respective frame. Then, we aggregate these foundation model-enhanced point clouds over time and project the resulting dense cloud into BEV space using average-pooling to generate BEV pseudo-labels as targets for distillation. Importantly, the point cloud is used only during training to generate the pseudo-labels and is not used during inference. This approach enables the online estimation of a neural BEV map with rich descriptive features, which improves detection and tracking beyond traditional BEV maps that utilize a predefined set of map classes.  We note that the strongest improvement is achieved when using DINOv2 as both (i) the backbone via ViT-Adapter-B and (ii) the source of features for BEV distillation, highlighting the synergy between backbone features and BEV supervision. We perform extensive experimental evaluations on the nuScenes~\cite{caesar2020nuscenes} and Argoverse~2~\cite{wilson2023argoverse} benchmarks to validate the effectiveness of our approach. The results demonstrate that our proposed \net achieves state-of-the-art performance in camera-based detection and tracking.

\begin{figure}[t]
    \centering
    \includegraphics[width=\linewidth]{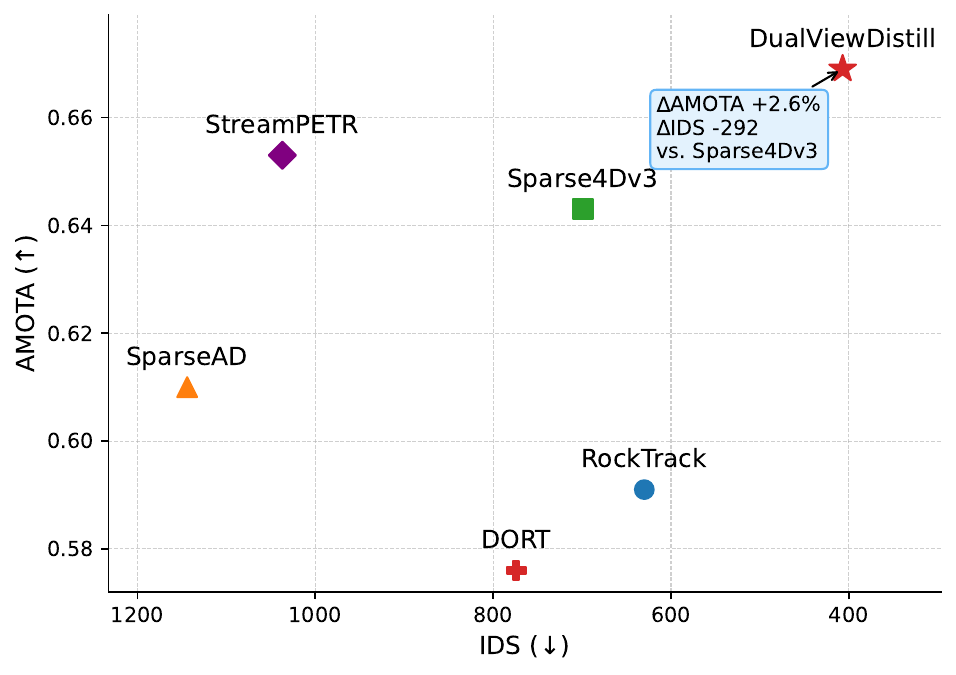}
    \caption{Previous SOTA methods vs. \net on the nuScenes 3D Multi-Object Tracking Benchmark. \net achieves superior performance across all key tracking metrics, notably improving AMOTA and reducing ID switches on the nuScenes test set.}
    \label{fig:cover_results}
\end{figure}

To summarize, our main contributions are as follows:
\begin{enumerate}[topsep=0pt]
    \item A dual-view detection and tracking framework that combines both PV and BEV camera image features, harnessing both object-centric and spatial reasoning advantages.
    \item An online estimated neural BEV map distilled from accumulated DINOv2 features, enabling a richer, more expressive representation of the scene beyond predefined BEV map classes, providing additional scene context to guide 3D object detection and tracking.
    \item Our experiments on the nuScenes benchmark validate the effectiveness of our approach, achieving state-of-the-art performance in camera-based detection and tracking, as illustrated in \cref{fig:cover_results}, where \net substantially improves AMOTA while reducing ID switches compared to previous methods.
    \item We further show that our performance gains generalize to the Argoverse~2 dataset, where we evaluate at near and long-range, observing consistent improvements over strong camera-only baselines and demonstrating robust long-range perception and generalization across datasets.
    \item In extensive ablation studies, we illustrate the influence of
various architectural and map supervision design choices.
    \item We make the code and pre-trained models publicly available at \mbox{\url{https://dualviewdistill.cs.uni-freiburg.de}}.
\end{enumerate}

\section{Related Work}
\label{sec:related-work}
In this section, we present an overview of prior work on camera-based 3D object detection, multi-object tracking, map-based perception, and the use of vision foundation models for perception in autonomous driving.

%%%%%%%%%%%%%%%%%%%%%%%%%%%%%%%%%%%%%%%%%%%%%%%%%%%%%%%%%%%%%%%%%%%%%%%%%%%%%%%%

\subsection{3D Object Detection}
Camera-based 3D object detection aims to predict 3D bounding boxes from multi-view images, addressing challenges such as depth estimation, multi-view fusion, and temporal fusion. Existing methods typically fall into two categories, dense BEV-based~\cite{li2022bevformer, yang2023bevformer, huang2021bevdet, huang2022bevdet4d, han2024exploring, li2023bevdepth, huang2022bevpoolv2, li2024bevnext} and sparse query-based~\cite{liu2022petr, jiang2024far3d, wang2023exploring, lin2022sparse4dv1, lin2023sparse4dv2, lin2023sparse4d, sun2024sparsedrive} approaches.

Dense BEV-based methods transform multi-view images into a unified BEV representation and perform object detection on the resulting BEV feature map. BEVFormer~\cite{li2022bevformer,yang2023bevformer} introduces learnable queries for each BEV feature grid cell, which are projected into camera views using the camera calibration to sample image features. These sampled features are then fused into BEV queries via deformable attention. An alternative BEV-based approach is Lift-Splat-Shoot (LSS)~\cite{philion2020lift}, which computes the outer product between an estimated per-pixel depth distribution and image features to lift them into a frustum-shaped point cloud. The resulting frustum point clouds from all cameras are merged and pooled into BEV, yielding a dense spatial representation. Building on LSS, BEVDet~\cite{huang2021bevdet} was the first method that applied this mechanism to camera-based 3D object detection. BEVDet4D~\cite{huang2022bevdet4d} further enhances temporal modeling by aligning BEV features across time via ego-motion compensation and concatenation, whereas HENet~\cite{xia2024henet} performs temporal fusion by attending to BEV features of adjacent time frames. BEVDepth~\cite{li2023bevdepth} improves depth estimation by supervising the depth network with LiDAR depth. BEVPoolv2~\cite{huang2022bevpoolv2} proposes a highly efficient GPU implementation of the LSS lifting operation to improve training and inference efficiency. Finally, BEVNeXt~\cite{li2024bevnext} introduces long-term temporal BEV fusion with an extended receptive field and employs a two-stage object decoder that refines instance-level BEV features using image features.

Sparse query-based methods use sparse 3D object queries to aggregate image features via global cross-attention~\cite{liu2022petr} or by projecting reference points onto camera views to sample image features~\cite{lin2022sparse4dv1,jiang2024far3d}, without relying on dense BEV representations. StreamPETR~\cite{wang2023exploring} extends this paradigm by propagating object queries to the next frame, storing them in a memory queue, and taking ego-motion into account during propagation to enhance temporal fusion. Far3D~\cite{jiang2024far3d} leverages a 2D object detector and a depth network to initialize 3D object queries, which are refined in the decoder, enabling far-range object detection. Far3D, RayDN~\cite{liu2024ray}, and Sparse4Dv3~\cite{lin2023sparse4d} introduce instance denoising where the network is tasked to denoise noisy ground truth boxes, which reduces false positives along the depth and improves convergence. The Sparse4D family~\cite{lin2022sparse4dv1, lin2023sparse4dv2, lin2023sparse4d} represents a progression of increasingly effective query-based designs. Sparse4Dv1 projects 3D anchor box keypoints into camera views to bilinearly sample multi-scale image features, which are then fused via deformable aggregation for 3D anchor box refinement. Sparse4Dv2 improves temporal modeling by propagating object queries to subsequent frames, similar to StreamPETR, and stores object queries in a memory queue for temporal cross-attention, while also leveraging LiDAR depth supervision to speed up convergence. Sparse4Dv3~\cite{lin2023sparse4d} introduces instance denoising, box confidence estimation via quality estimation, and an end-to-end tracking extension. This progression results in strong performance across both detection and multi-object tracking benchmarks.

Our approach builds upon BEVNeXt and Sparse4Dv3 but introduces several key improvements. Unlike prior methods, which rely exclusively on either PV or BEV features, we integrate both representations in a unified manner, without requiring any extra head, thereby leveraging their complementary strengths. Additionally, existing methods do not incorporate latent BEV maps guided by foundation model features, which can provide strong cues and spatial reasoning for detection and tracking.
Specifically, such maps provide structured spatial priors for object placement (e.g., vehicles on roads) and motion patterns (e.g., vehicles following lanes), which enhance tracking.

\subsection{3D Multi-Object Tracking}
Camera-based 3D multi-object tracking methods can be categorized into tracking-by-detection and end-to-end query-based approaches. Tracking-by-detection methods rely on explicit association and motion modeling to maintain object identities across frames. StreamPETR~\cite{wang2023exploring} adopts the CenterPoint tracking~\cite{yin2021center} paradigm, where detections are greedily matched to existing tracks based on their ground-plane centers. Estimated object velocities are used both to predict forward displacements during matching and to propagate unmatched tracks. ByteTrackV2~\cite{Zhang2023ByteTrackV22A} addresses missed detections caused by noisy detection scores with a two-stage data association strategy that recovers tracklets from low-score boxes, reducing fragmented trajectories. RockTrack~\cite{li2024rocktrack} fuses motion and visual cues in a joint association module, using motion and image observations together with an appearance similarity metric to capture inter-object affinity and reduce mismatches.

Query-based approaches instead propagate object queries directly frame-by-frame, modeling temporal associations and performing detection and tracking jointly, end-to-end, in a single stage~\cite{zeng2022motr, yu2023motrv3, meinhardt2022trackformer, zhang2022mutr3d, pang2023standing}. MUTR3D~\cite{zhang2022mutr3d} and PF-Track~\cite{pang2023standing} extend this paradigm to 3D tracking. Sparse4Dv3~\cite{lin2023sparse4d} introduces a simple yet effective query-based approach: previous-frame detection queries are reused in the next frame, and objects retain their track IDs if they remain confident detections. To further enhance this query propagation, noisy instances from the last frame are projected to the current frame and denoised during training. Compared to tracking-by-detection, query-based methods avoid complex post-processing steps such as explicit object association and state prediction.

%%%%%%%%%%%%%%%%%%%%%%%%%%%%%%%%%%%%%%%%%%%%%%%%%%%%%%%%%%%%%%%%%%%%%%%%%%%%%%%%

\subsection{Map-Based Perception}
Map-based perception leverages road geometry and semantic priors from maps to improve 3D object detection. LiDAR-based methods such as HDNet~\cite{yang2018hdnet}, MapFusion~\cite{fang2021mapfusion}, and MENet~\cite{huang2023menet} fuse rasterized HD maps with LiDAR features in BEV, while LaneFusion~\cite{fujimoto2022lanefusion} incorporates lane direction to refine heading estimates. Despite their benefits, these approaches depend mostly on hand-annotated offline maps that are not always available and have seen limited use in camera-only settings. Recent works extend map-based perception beyond LiDAR. BEVMap~\cite{chang2024bevmap} projects hand-annotated HD maps into camera views to guide LSS-based depth estimation for camera-only object detection, and Neural Map Prior~\cite{xiong2023neural} accumulates latent BEV features into a global neural map that serves as an online prior for online map estimation. However, most approaches (i) target LiDAR detection, (ii) assume fixed semantic map classes, and (iii) are not integrated with modern transformer-based detectors.

In contrast, we propose foundation-model-guided BEV maps distilled from DINOv2 features, providing rich, online, up-to-date priors without requiring manually labeled HD maps. Such BEV maps can integrate seamlessly with modern transformer-based architectures, such as our camera-only joint detector and tracker.

%%%%%%%%%%%%%%%%%%%%%%%%%%%%%%%%%%%%%%%%%%%%%%%%%%%%%%%%%%%%%%%%%%%%%%%%%%%%%%%%

\subsection{Vision Foundation Models for Scene Understanding}

Vision foundation models, such as CLIP~\cite{radford2021learning}, Stable Diffusion~\cite{rombach2022high}, and DINOv2~\cite{oquab2023dinov2}, are typically trained on large-scale datasets using self-supervised learning, allowing them to learn general-purpose features that can be effectively adapted to diverse vision tasks. DINOv2, trained only on raw images in a self-supervised manner, yields descriptive general-purpose image features that enable image classification, semantic segmentation, and depth estimation via linear probing. Recent works such as SPINO~\cite{kappeler2024few} and PASTEL~\cite{voedisch2025pastel} leverage frozen DINOv2 features to enable panoptic segmentation with only 10 annotated images. Subsequent work also leverages DINOv2 features for LiDAR~\cite{hindel2025label} and multispectral semantic segmentation~\cite{hurtado2025hyperspectral}. In the context of BEV-based perception, several methods~\cite{schramm2024bevcar, barin2024robust, gosala2024letsmap} adapt a frozen DINOv2 as the image backbone to improve robustness to adverse weather conditions. However, these methods primarily use DINOv2 as the image backbone rather than leveraging it to learn DINOv2 features in BEV space. 

Beyond using foundation models as the backbone, several works focus on lifting their features into 3D or BEV space for enhanced scene understanding. Lexicon3D~\cite{man2024lexicon3d} probes and evaluates various vision foundation models for 3D scene understanding across different semantic and geometric 3D tasks, showing that DINOv2 achieves the best overall performance. OccFeat~\cite{sirko2024occfeat} pretrains a BEV model via occupancy prediction and DINOv2 feature distillation in a self-supervised manner to enable label-efficient BEV map segmentation. DistillNeRF~\cite{wang2024distillnerf} further explores distillation by transferring per-scene optimized NeRFs and DINOv2 features into an estimated 3D voxel representation via differentiable rendering, enabling zero-shot 3D semantic occupancy prediction. Nevertheless, due to the high computational and memory costs associated with their 3D voxel-based model, this approach remains impractical for 3D object detection, which requires high image resolution. Instead, our method pools feature point clouds into BEV before distillation, maintaining efficiency while leveraging rich DINOv2 features.

Unlike prior methods, which either use foundation models only as backbones or for occupancy prediction, we explicitly distill their learned features into BEV space to enhance detection and tracking performance. To the best of our knowledge, our work is the first to leverage foundation model guided BEV maps for camera-only joint 3D object detection and multi-object tracking.
\section{Technical Approach}
\label{sec:method}

\begin{figure*}[t]
    \centering
    \includegraphics[width=\linewidth]{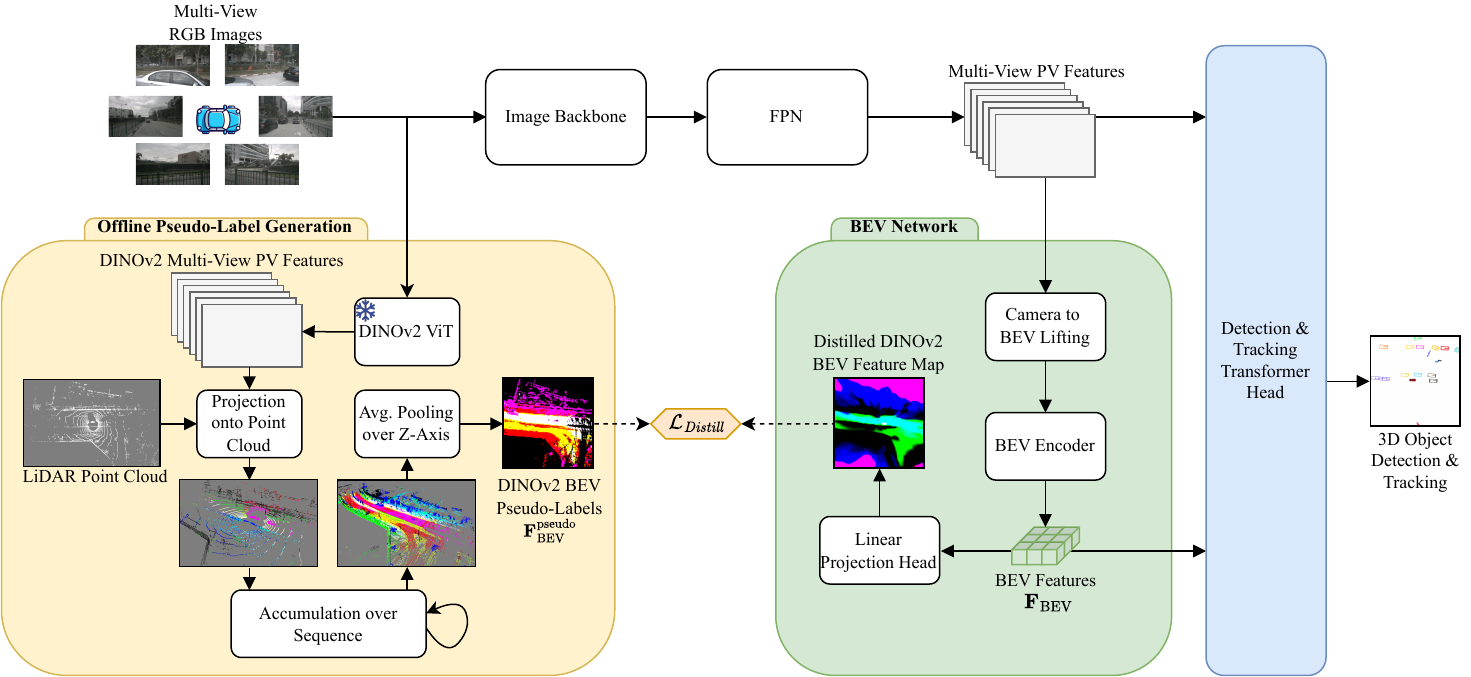}
    \caption{
    Overview of our proposed \net approach. We jointly leverage PV and BEV camera features, while enriching the BEV representation through DINOv2-guided distillation. Pseudo-labels for distillation are generated by projecting DINOv2 features into BEV space via the LiDAR point cloud before training. Both components jointly improve 3D object detection and tracking.}
    \label{fig:overview}
\end{figure*}

We propose \net, a hybrid camera-based 3D object detection and tracking framework that integrates both perspective-view (PV) and bird's-eye-view (BEV) features, while further leveraging foundation model knowledge through BEV map distillation. Built on a transformer-based detection and tracking architecture~\cite{lin2023sparse4d}, we introduce a novel BEV distillation module guided by DINOv2~\cite{oquab2023dinov2} features and joint PV--BEV deformable aggregation. An overview of our proposed \net architecture is shown in Fig.~\ref{fig:overview}.

%%%%%%%%%%%%%%%%%%%%%%%%%%%%%%%%%%%%%%%%%%%%%%%%%%%%%%%%%%%%%%%%%%%%%%%%%%%%%%%%

\subsection{Network Architecture}
\label{ssec:method-architecture}

Our network takes multi-view RGB images from surround-view cameras as input and outputs 3D bounding boxes and object tracks. The architecture consists of four components:

\subsubsection{Image Backbone}

We employ standard image backbones and a feature pyramid network (FPN) to extract multi-scale, multi-view PV features for both the BEV network and the detection head. In addition, we use a frozen DINOv2 ViT-S model in conjunction with a multi-scale ensemble~\cite{voedisch2025pastel} to generate fine-grained DINOv2-based BEV pseudo-labels for distillation. The multi-scale ensemble is particularly important as the native DINOv2 ViT patch size (i.e., 14) yields relatively coarse feature maps. Multi-scale aggregation provides fine spatial detail for dense BEV supervision.

\subsubsection{BEV Network} 

To complement PV features with spatially structured BEV representations, we utilize a BEV network that lifts PV features to the top-down BEV metric space. We adopt the Lift-Splat-Shoot (LSS)~\cite{philion2020lift, li2024bevnext} mechanism. Concretely, given per-view image features $\mathbf{F}_{\text{PV}}(u,v)\in\mathbb{R}^C$ and an estimated normalized discrete depth probability vector $\mathbf{P}_{\text{depth}}(u,v)\in[0,1]^D$ (s.t. $\sum_{d=1}^{D}\mathbf{P}_{\text{depth}}(u,v,d)=1$ for every $(u,v)$), LSS forms a frustum feature cloud by taking the outer product:
\begin{equation}
\mathbf{F}_{\text{frustum}}(u,v) = \mathbf{P}_{\text{depth}}(u,v) \otimes \mathbf{F}_{\text{PV}}(u,v)\in\mathbb{R}^{D \times C},
\end{equation}
and unprojects $(u,v,d)$ to 3D using the camera intrinsics and extrinsics to obtain per-camera 3D points with attached features. Feature point clouds from all cameras are then merged and summed into BEV grid cells to obtain $\mathbf{F}_{\text{BEV}}$. The depth map is predicted by a DepthNet, which is supervised with LiDAR depth during training. We attach a Conditional Random Field (CRF) modulation~\cite{li2024bevnext} to the depth estimator: a CRF refines $\mathbf{P}_{\text{depth}}$ by enforcing color-consistent smoothness in the depth map using the RGB image. This CRF-guided refinement encourages object-level depth consistency before lifting and reduces lifting artifacts. The BEV features are finally processed with a BEV encoder that consists of multiple ResNet blocks and a small FPN.

\begin{figure}
    \centering
    \includegraphics[width=1.0\linewidth]{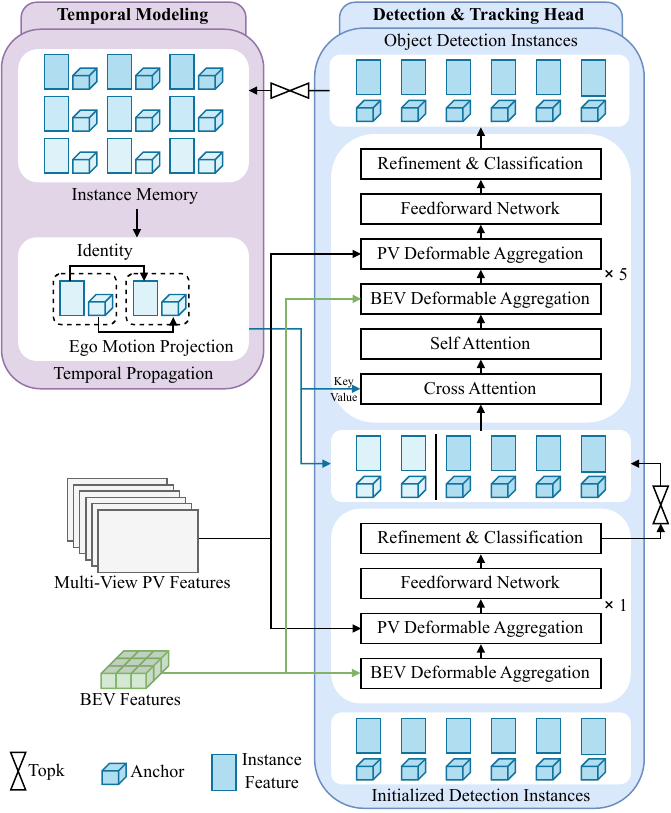}
    \caption{Detection and Tracking Transformer Head. Object queries consisting of anchors and instance features interact via deformable aggregation with PV and BEV features. An instance memory is used for temporal fusion and to propagate queries for tracking.}
    \label{fig:head}
\end{figure}

\subsubsection{Detection and Tracking Head}

The detection and tracking head shown in Fig.~\ref{fig:head} follows a transformer architecture. It processes learnable object queries consisting of 3D anchor boxes and instance features. 
Let $A\in\mathbb{R}^{M\times 11}$ and $F\in\mathbb{R}^{M\times C}$ denote the anchors and instance features for $M$ objects with feature channel size $C$. Each anchor box is parameterized with its center location, dimension, yaw angle, and velocity:

{\small
\begin{equation}
a_i = (x,y,z,\ln w, \ln h, \ln l,\sin yaw, \cos yaw, v_x,v_y,v_z).
\end{equation}}

Anchor centers are initialized by $k$-means clustering over the training set. Each anchor box is encoded into a high-dimensional embedding using an anchor encoder and then fused with the instance feature through addition before each transformer block to form object queries. Object queries are then iteratively refined in each decoder block.
The decoder network is composed of a non-temporal block followed by a stack of five temporal blocks, as shown in Fig.~\ref{fig:head}. The \textit{non-temporal block} consists of BEV Deformable Aggregation, PV Deformable Aggregation~\cite{lin2023sparse4dv2}, a feedforward network (FFN), and 3D box refinement/classification heads. Each \textit{temporal block} consists of temporal cross-attention (attending to queries from the last frame), self-attention across current queries, BEV and PV deformable aggregation, an FFN, and refinement/classification heads. The refinement head includes a regression layer that predicts residuals from the current anchor to the ground truth, thereby refining the box iteratively.

\paragraph*{PV and BEV Deformable Aggregation} In each decoder layer, object queries interact with both PV and BEV features using deformable aggregation. For PV aggregation, we generate multiple 3D anchor keypoints for each query's anchor to cover the object, comprising fixed keypoints (center and face centers of the box) and learnable keypoints predicted from the instance feature. Keypoints are projected into the image plane of each camera using known intrinsics and extrinsics parameters to bilinearly sample multi-scale, multi-view PV features \(\mathbf{F}_{\text{PV}}\). For BEV aggregation, the same 3D anchor keypoints are projected into the BEV feature grid to bilinear-sample features \(\mathbf{F}_{\text{BEV}}\) accordingly. The sampled multi-keypoint, multi-view, multi-scale features are then fused via learned attention weights and used to update the query's instance feature with residual connections:
\begin{equation}
    \begin{aligned}
        \mathbf{f}_i'  &= \mathbf{f}_i + \text{BEV-DeformAgg}(\mathbf{f}_i, \mathbf{a}_i, \mathbf{F}_{\text{BEV}}), \\
        \mathbf{f}_i'' &= \mathbf{f}_i' + \text{PV-DeformAgg}(\mathbf{f}_i', \mathbf{a}_i, \mathbf{F}_{\text{PV}}),
    \end{aligned}
\end{equation}
where \(\mathbf{f}_i\) is the instance feature and \(\mathbf{a}_i\) the anchor of the \(i\)-th object query while \(\text{DeformAgg}_{\cdot}(\cdot)\) denotes deformable aggregation from the respective modality. This novel sequential aggregation enables effective fusion of both BEV and PV representations, leveraging their complementary strengths.

\subsubsection{Temporal Modeling and Tracking}
For tracking, we maintain a temporal memory of instance features and anchors from the previous frame and propagate them to the current frame. Instance features are copied unchanged, while anchor boxes are motion-propagated by first applying the last-frame velocity estimate to the box and then transforming from the previous ego pose to the current ego pose. The propagated anchor boxes are re-encoded by the anchor encoder to obtain current-frame anchor embeddings, which are fused through addition with the propagated instance features to obtain object queries. We then combine these propagated queries with the top-\(k\) high-confident detection queries from the non-temporal block to form input queries for the temporal blocks. Tracking is realized by persistent IDs: if a propagated query (same internal ID as last frame) yields a confident detection in the current frame, the same ID is retained; newly emerged high-confidence queries are assigned fresh IDs. Temporal blocks perform cross-attention to the propagated queries from the last frame by using their object queries as the temporal keys and values.

%%%%%%%%%%%%%%%%%%%%%%%%%%%%%%%%%%%%%%%%%%%%%%%%%%%%%%%%%%%%%%%%%%%%%%%%%%%%%%%%

\subsection{Foundation Model Guided BEV Map Distillation}
\label{ssec:method-distillation}

To further enrich the BEV representation with descriptive features from pretrained foundation models, we distill DINOv2 features into BEV features during training via offline pseudo-label generation. Beyond the sparse, instance-level supervision from the detection head, DINOv2 to BEV distillation provides dense, per-cell supervision directly in BEV, improving coverage across the BEV grid and complementing box-level signals.

\subsubsection{Offline Pseudo-Label Generation}
We extract fine-grained DINOv2 features using a multi-scale ensemble for each surround-view camera. These features are projected onto the LiDAR point cloud using known camera intrinsics and extrinsics. For each visible LiDAR point \(\mathbf{p} \in \mathbb{R}^3\), we sample image features from all views that observe \(\mathbf{p}\) and assign the average DINOv2 feature as \(\mathbf{f}_\text{DINO}(\mathbf{p})\). To generate dense supervision and obtain supervision for occluded regions, we accumulate LiDAR point clouds over the entire sequence to create a dense, static point cloud. We also aggregate points per object in object-centric coordinates using ground truth bounding boxes and tracks. These point sets are voxelized and then averaged along the height (Z-axis) to obtain a dense 2D BEV feature grid:
\begin{equation}
    \mathbf{F}_\text{BEV}^{\text{pseudo}}(x, y) = \frac{1}{|\mathcal{V}_{x,y}|} \sum_{\mathbf{p}_i \in \mathcal{V}_{x,y}} \mathbf{f}_\text{DINO}(\mathbf{p}_i),
\end{equation}
where \(\mathcal{V}_{x,y}\) is the set of voxelized 3D points falling into BEV cell \((x, y)\). The resulting \(\mathbf{F}_\text{BEV}^{\text{pseudo}}\) pseudo-labels are used for supervision during training. We rely on LiDAR for lifting because it provides the most accurate 3D geometry. Our oracle analysis (\cref{ssec:result-abl-oracle}) shows its effectiveness and that DINOv2 features further boost both detection and tracking. LiDAR is used only to generate depth ground truths and pseudo-labels during training, and no LiDAR is used at inference.

\subsubsection{Distillation Loss}

During training, we add a linear projection head \(g(\cdot)\) to map our BEV features to the DINOv2 feature space and minimize the cosine similarity loss:

{\small
\begin{equation}
    \mathcal{L}_{\text{distill}} = \frac{1}{|\Omega|} \sum_{(x,y) \in \Omega} \left(1 - \text{sim}_{\cos}\left(g(\mathbf{F}_{\text{BEV}}(x,y)), \mathbf{F}_{\text{BEV}}^{\text{pseudo}}(x,y)\right)\right),
\end{equation}}
where \(\Omega\) is the set of BEV cells with valid LiDAR-projected DINOv2 features. Supervising only such cells avoids imprinting LiDAR sampling artifacts onto the camera BEV features (see ablation in Tab.~\ref{tab:ablation-dinov2-sup}). This enables the BEV encoder to learn semantic and geometric priors from rich DINOv2 representations while maintaining end-to-end detection and tracking supervision.

%%%%%%%%%%%%%%%%%%%%%%%%%%%%%%%%%%%%%%%%%%%%%%%%%%%%%%%%%%%%%%%%%%%%%%%%%%%%%%%%

\subsection{Training and Optimization}
\label{ssec:method-training}

The total training loss is defined as:
\begin{equation}
\mathcal{L}_{\text{total}} = \lambda_{\text{det}} \mathcal{L}_{\text{det}} + \lambda_{\text{distill}} \mathcal{L}_{\text{distill}} + \lambda_{\text{depth}} \mathcal{L}_{\text{depth}},
\end{equation}
where \(\mathcal{L}_{\text{det}}\) includes classification (focal loss), bounding box regression (L1 loss), yawness (cross-entropy loss), and centerness (focal loss)~\cite{lin2023sparse4d}. \(\mathcal{L}_{\text{depth}}\) supervises both the LSS depth estimator (binary cross-entropy) and an auxiliary depth head on PV features (L1 loss). \(\mathcal{L}_{\text{distill}}\) is the DINOv2 distillation loss described above. We empirically set the loss weights \(\lambda_{\text{det}}\),  \(\lambda_{\text{distill}}\), and  \(\lambda_{\text{depth}}\) to balance training dynamics.

\section{Experimental Evaluation}
\label{sec:result}
In this section, we first present the benchmarking results of \net for camera-only 3D object detection and multi-object tracking on the nuScenes~\cite{caesar2020nuscenes} and Argoverse~2~\cite{wilson2023argoverse} datasets, with multiple backbones and input resolutions in \cref{tab:results-3d-val,tab:results-3d-test,tab:results-track-val,tab:results-track-test,tab:results-argoverse2-150m,tab:results-argoverse2-50m}. We then present a series of ablation results to study the following questions: 
\begin{itemize}
    \item Does a hybrid model that leverages PV and BEV features outperform using either view alone?
    \item Do foundation model guided BEV maps provide complementary semantic and geometric information that translates into measurable gains in detection and tracking?
    \item Which design choices in the BEV network, pseudo-label generation, and distillation most strongly affect performance?
\end{itemize}
The ablation results presented in \cref{tab:ablation-bev-map,tab:ablation-dinov2-gen,tab:ablation-dinov2-sup,tab:ablation-dinov2-loss} show that combining PV and BEV yields the best performance and that DINOv2 BEV distillation consistently improves both NDS/mAP and AMOTA scores. We further analyze robustness under adverse weather and illumination in \cref{tab:analysis-weather}. Finally, we complement these quantitative findings with qualitative visualizations of detection and tracking on challenging scenes in \cref{fig:qualitative-nuscenes-test,fig:qualitative-nuscenes-validation,fig:qualitative-argoverse2-validation,fig:qualitative-nuscenes-validation-tracking}, highlighting fewer ID switches and more precise object localization.

\begin{table*}
\footnotesize
\centering
\caption{Results of 3D object detection on the nuScenes validation set.}
\label{tab:results-3d-val}
\setlength\tabcolsep{1.7pt}
\begin{threeparttable}
    \begin{tabular}{l | c c | c c | c c c c c}
        \toprule
        \textbf{Method} & \textbf{Backbone} & \textbf{Image Size} & \textbf{NDS$\uparrow$} & \textbf{mAP$\uparrow$} & \textbf{mATE$\downarrow$} & \textbf{mASE$\downarrow$} & \textbf{mAOE$\downarrow$} & \textbf{mAVE$\downarrow$} & \textbf{mAAE$\downarrow$} \\
        \midrule
        BEVDepth~\cite{li2023bevdepth} & ResNet101 & $512 \times 1408$ & 0.535 & 0.412 & 0.565 & 0.266 & 0.358 & 0.331 & 0.190 \\ % from Sparse4Dv3, BEVNeXt, StreamPETR, BEVDepth
        SOLOFusion~\cite{Park2022TimeWT} & ResNet101 & $512 \times 1408$ & 0.582 & 0.483 & 0.503 & 0.264 & 0.381 & 0.246 & 0.207 \\ % from SOLOFusion, Sparse4Dv3, BEVNeXt
        SparseBEV\textsuperscript{\textdagger}~\cite{liu2023sparsebev} & ResNet101 & $512 \times 1408$ & 0.592 & 0.501 & 0.562 & 0.265 & 0.321 & 0.243 & 0.195 \\ % from SparseBEV paper
        StreamPETR\textsuperscript{\textdagger}~\cite{wang2023exploring} & ResNet101 & $512 \times 1408$ & 0.592 & 0.504 & 0.569 & 0.262 & 0.315 & 0.257 & 0.199 \\ % % from StreamPETR paper
        Sparse4Dv2\textsuperscript{\textdagger}\cite{lin2023sparse4dv2} & ResNet101 & $512 \times 1408$ & 0.594 & 0.506 & 0.540 & 0.258 & 0.348 & 0.239 & 0.184 \\
        Far3D\textsuperscript{\textdagger}\cite{jiang2024far3d} & ResNet101 & $512 \times 1408$ & 0.594 & 0.510 & 0.551 & 0.258 & 0.372 & 0.238 & 0.195 \\
        BEVNeXt\textsuperscript{\textdagger}~\cite{li2024bevnext} & ResNet101 & $512 \times 1408$ & 0.597 & 0.500 & \textbf{0.487} & 0.260 & 0.343 & 0.245 & 0.197 \\ % BEVNeXT paper
        RayDN\textsuperscript{\textdagger}~\cite{liu2024ray} & ResNet101 & $512 \times 1408$ & 0.604  & 0.518 & 0.541 & 0.260 & 0.315 & 0.236 & 0.200 \\
        Sparse4Dv3\textsuperscript{\textdagger}~\cite{lin2023sparse4d} & ResNet101 & $512 \times 1408$ & 0.623 & \textbf{0.537} & 0.511 & 0.255 & 0.306 & 0.194 & 0.192 \\ % from Sparse4Dv3 paper
        \rowcolor{Gray} \net \textsuperscript{\textdagger} & ResNet101 & $512 \times 1408$ & \textbf{0.625} & 0.532 & 0.492 & \textbf{0.253} & \textbf{0.299} & \textbf{0.181} & \textbf{0.190} \\
        \midrule
        StreamPETR\textsuperscript{\textdaggerdbl}~\cite{wang2023exploring} & ViT-Adapter-B & $256 \times 704$ & 0.566 & 0.467 & 0.611 & 0.270 & 0.329 & 0.269 & 0.196 \\ % own trained with StreamPETR code
        Sparse4Dv3\textsuperscript{\textdaggerdbl}~\cite{lin2023sparse4d} & ViT-Adapter-B & $256 \times 704$ & 0.597 & 0.493 & 0.517 & \textbf{0.259} & 0.333 & 0.215 & \textbf{0.171} \\
        \rowcolor{Gray} \net & ViT-Adapter-B & $256 \times 704$ & \textbf{0.619} & \textbf{0.527} & \textbf{0.503} & \textbf{0.259} & \textbf{0.311} & \textbf{0.194} & 0.180 \\
        \midrule
        SRCN3D~\cite{shi2022srcn3d} & VoVNet-99 & $640 \times 1600$ & 0.475 & 0.396 & 0.737 & 0.294 & 0.278 & 0.728 & 0.197 \\
        SparseAD~\cite{zhang2024sparsead} & VoVNet-99 & $640 \times 1600$ & 0.578 & 0.475 & 0.556 & - & 0.295 & 0.288 & - \\
        HENet~\cite{xia2024henet} & VoVNet-99 & $640 \times 1152$ & 0.599 & 0.502 & 0.465 & 0.261 & 0.335 & 0.267 & 0.197 \\
        Sparse4Dv3\textsuperscript{\textdaggerdbl}~\cite{lin2023sparse4d} & VoVNet-99 & $640 \times 1600$ & 0.649 & 0.564 & 0.472 & 0.259 & 0.213 & 0.190 & 0.189 \\
        \rowcolor{Gray} \net\textsuperscript{\textdagger} & VoVNet-99 & $640 \times 1600$ & \textbf{0.669} & \textbf{0.588} & \textbf{0.440} & \textbf{0.245} & \textbf{0.196} & \textbf{0.184} & \textbf{0.184} \\
        \bottomrule
    \end{tabular}
    \footnotesize
    \textdagger:~Uses pre-trained weights from the nuImage dataset.
    \textdaggerdbl:~Baselines trained with the code provided by the authors.
\end{threeparttable}
\end{table*}

\begin{table*}
\footnotesize
\centering
\caption{Results of 3D object detection on the nuScenes test set.}
\label{tab:results-3d-test}
\setlength\tabcolsep{2.4pt}
\begin{threeparttable}
    \begin{tabular}{l | c c | c c | c c c c c}
        \toprule
        \textbf{Method} & \textbf{Backbone} & \textbf{Image Size} & \textbf{NDS$\uparrow$} & \textbf{mAP$\uparrow$} & \textbf{mATE$\downarrow$} & \textbf{mASE$\downarrow$} & \textbf{mAOE$\downarrow$} & \textbf{mAVE$\downarrow$} & \textbf{mAAE$\downarrow$} \\
        \midrule
        SparseAD~\cite{zhang2024sparsead} & ViT-L & $800 \times 1600$ & 0.657 & 0.592 & 0.494 & 0.243 & 0.286 & 0.241 & 0.132 \\
        StreamPETR~\cite{wang2023exploring} & ViT-L~\cite{fang2024eva} & $800 \times 1600$ & 0.676 & 0.620 & 0.470 & 0.241 & 0.\textbf{258} & 0.236 & 0.134 \\ % from StreamPETR paper
        RayDN~\cite{liu2024ray} & ViT-L & $800 \times 1600$ & 0.686 & 0.631 & 0.437 & 0.235 & 0.283 & 0.220 & \textbf{0.120} \\ % from nuScenes leaderboard
        Far3D~\cite{jiang2024far3d} & ViT-L & $1536 \times 1536$ & 0.687 & \textbf{0.635} & 0.432 & 0.237 & 0.278 & 0.227 & 0.130 \\ % from Far3D paper
        Sparse4Dv3~\cite{lin2023sparse4d} & ViT-L~\cite{fang2024eva} & $640 \times 1600$ & 0.694 & 0.630 & 0.379 & 0.235 & 0.281 & 0.184 & 0.127 \\
        \rowcolor{Gray} \net & ViT-L~\cite{fang2024eva} & $640 \times 1600$  & \textbf{0.695} & 0.621 & \textbf{0.360} & \textbf{0.227} & 0.272 & \textbf{0.168} & 0.126 \\
        \bottomrule
    \end{tabular}
    \footnotesize
\end{threeparttable}
\end{table*}

%%%%%%%%%%%%%%%%%%%%%%%%%%%%%%%%%%%%%%%%%%%%%%%%%%%%%%%%%%%%%%%%%%%%%%%%%%%%%%%%

\subsection{Implementation and Training Details}

\subsubsection*{Detection and Tracking Head}
Our method builds upon Sparse4Dv3~\cite{lin2023sparse4d} for the detection and tracking head, using multi-scale image features from scales 1/4, 1/8, 1/16, and 1/32. Following Sparse4Dv2~\cite{lin2023sparse4dv2} and Sparse4Dv3~\cite{lin2023sparse4d}, our head employs a 6-layer decoder with 900 object queries and $N_t=600$ temporal queries, with an embedding dimension of 256. Instance denoising is enabled unless stated otherwise.

\subsubsection*{BEV Network and Lifting}
We adopt the BEVNeXt~\cite{li2024bevnext} architecture for the BEV network. For $256\times704$ input resolution, we use a BEV grid of $128\times128$ and scale 1/8 for the image feature to BEV lifting. For higher resolutions, we set the BEV grid to $256\times256$ and use image features of scale 1/16 for the BEV lifting.

\subsubsection*{Backbones and Pretraining}
We evaluate our method across different backbones, including ResNet-101~\cite{he2016deep}, ViT-Adapter-B~\cite{chen2023vision}, VoVNet-99~\cite{lee2019energy}, and ViT-L~\cite{fang2024eva}. The ResNet-101 is initialized with weights pretrained on nuImages, the ViT-Adapter-B is pretrained with DINOv2~\cite{oquab2023dinov2}, VoVNet-99 is pretrained with DD3D~\cite{park2021pseudo}, and ViT-L is initialized with EVA-02~\cite{fang2024eva} weights and further pretrained on COCO~\cite{lin2014microsoft} and Objects365~\cite{shao2019objects365}, following prior works~\cite{wang2023exploring,lin2023sparse4d,jiang2024far3d}. The FPN is not used for the ViT-Adapter-B backbone, since ViT-Adapter already outputs multi-scale features.

\subsubsection*{Optimization and Training}
We train the models for 100 epochs, depending on the backbone, using AdamW. We employ cosine learning rate decay with a 500-iteration warm-up and set the base learning rate to 2e-4. We use a batch size of 32 for ResNet-101 and ViT-Adapter-B, and 24 for VoVNet-99 and ViT-L. We follow a sequential iteration training strategy, processing one frame per step with a temporal instance memory from previous frames. Only keyframes, i.e., frames with ground truth annotations provided at $\SI{2}{\hertz}$, are used during both training and inference. To stabilize training, we apply gradient clipping with a maximum norm of 10 for the ViT-Adapter-B backbone and 5 for all other backbones. Additionally, for all models except ViT-Adapter-B, we use a reduced learning rate for the backbone, set to 0.1$\times$ of the standard learning rate. For ViT-L, we further scale the BEV network learning rate by 0.5$\times$. We do not use class-balanced Grouping and Sampling (CBGS)~\cite{zhu2019class}.

\subsubsection*{Data Augmentations and Auxiliary Supervision}
We apply both image-level and 3D/BEV-level augmentations~\cite{huang2021bevdet}. BEV/3D augmentations consist of scaled rigid transformations that rotate, scale, and flip the BEV grid together with the ground truth boxes. Camera extrinsics are updated accordingly to preserve correct projections. In addition, we use dense depth supervision as an auxiliary task for both LSS and PV-based depth heads.

\subsubsection*{Implementation Details for Argoverse~2}
To handle the $\SI{150}{\meter}$ perception range and the larger set of object categories (26 classes vs. 10 on nuScenes), we increase the number of object queries from 900 to 1800 and temporal queries from $N_t=600$ to $N_t=1200$. This provides a sufficient number of queries for the larger perception range and stabilizes training. The BEV network uses a $256\times256$ BEV grid and lifts image features of scale 1/16 to BEV. The perception range during training is set to $\SI{152.4}{\meter}\times\SI{152.4}{\meter}$ following Far3D~\cite{jiang2024far3d}. We train for 40 epochs with batch size 24 on the train split and adopt a VoVNet-99 backbone initialized with FCOS3D weights pretrained on nuScenes (in line with prior works~\cite{jiang2024far3d, liu2024ray}). Inputs use a $960\times640$ image resolution. Since the front-view image in Argoverse~2 has a different resolution than the other views, we first resize it to a consistent resolution across cameras and then apply the same image augmentations as for the remaining views. To reduce memory usage, we keep the BEV grid range at $\SI{50}{\meter}$ while allowing the head to exploit PV features beyond the BEV extent, thereby retaining an effectively unbounded perception range. All other model and training parameters match our nuScenes setup.

%%%%%%%%%%%%%%%%%%%%%%%%%%%%%%%%%%%%%%%%%%%%%%%%%%%%%%%%%%%%%%%%%%%%%%%%%%%%%%%%

\subsection{Datasets and Metrics}
We evaluate our approach on both the nuScenes~\cite{caesar2020nuscenes} and Argoverse~2~\cite{wilson2023argoverse} benchmarks to assess our model's performance.

\textit{nuScenes~\cite{caesar2020nuscenes}} is a large-scale autonomous driving dataset containing 1000 scenes collected in urban environments using 6 surround-view cameras. Each scene is approximately $\SI{20}{\second}$ long and annotated at $\SI{2}{\hertz}$. The official split assigns 700/150/150 scenes to train/val/test. In addition, a single 32-beam LiDAR operating at $\SI{20}{\hertz}$ is available for training supervision. We use the official camera-only setting without LiDAR input at inference time. Models are trained and evaluated on 10 object categories within a $\SI{50}{\meter}$ range. For 3D object detection, we follow the official evaluation protocol and report the nuScenes Detection Score (NDS) and mean Average Precision (mAP), as well as the five true-positive metrics: mean Average Translation Error (mATE), mean Average Scale Error (mASE), mean Average Orientation Error (mAOE), mean Average Velocity Error (mAVE), and mean Average Attribute Error (mAAE). For multi-object tracking, we also adopt the nuScenes tracking benchmark metrics and report Average Multi-Object Tracking Accuracy (AMOTA), Average Multi-Object Tracking Precision (AMOTP), and Identity Switches (IDS). For completeness, we also include Recall, Multi-Object Tracking Accuracy (MOTA), and Multi-Object Tracking Precision (MOTP). All results are reported on the official validation and test sets.

\textit{Argoverse~2~\cite{wilson2023argoverse}} is a large-scale dataset for perception and prediction in autonomous driving with 1000 scenes, each 15 seconds long and annotated at $\SI{10}{\hertz}$. The official split assigns 700/150/150 scenes to train/val/test, respectively. It provides seven high-resolution camera ring images with a combined \ang{360} field of view, and two 32-beam LiDAR scans at $\SI{10}{\hertz}$ are available for training supervision. We follow the official camera-only setting and never use LiDAR at inference. We evaluate 26 object categories at both $\SI{50}{\meter}$ and $\SI{150}{\meter}$ ranges to cover near- and long-range detection. In addition to mean Average Precision (mAP), we report the Composite Detection Score (CDS), which is the dataset's primary metric, along with the three true-positive metrics: mean Average Translation Error (mATE), mean Average Scale Error (mASE), and mean Average Orientation Error (mAOE).

\begin{table*}
\footnotesize
\centering
\caption{Results of 3D multi-object tracking on the nuScenes validation set.}
\label{tab:results-track-val}
\setlength\tabcolsep{2.5pt}
\begin{threeparttable}
    \begin{tabular}{l | c c | c c c c c c }
        \toprule
        \textbf{Method} & \textbf{Backbone} & \textbf{Image Size} & \textbf{AMOTA$\uparrow$} & \textbf{AMOTP$\downarrow$} & \textbf{IDS$\downarrow$} & \textbf{Recall$\uparrow$} & \textbf{MOTA$\uparrow$} & \textbf{MOTP$\downarrow$} \\
        \midrule
        ADA-Track~\cite{ding2024ada} & ResNet101 & $640 \times 1600$ & 0.392 & 1.375 & 897 & 0.523 & 0.363 & -  \\
        DQTrack~\cite{li2023end} & ResNet101 & $512 \times 1408$ & 0.407 & 1.318 & 1003 & - & - & - \\ % from Sparse4Dv3 paper
        Sparse4Dv3~\cite{lin2023sparse4d} & ResNet101 & $512 \times 1408$ & 0.567 & 1.027 & 557 & 0.658 & 0.515 & 0.621 \\
        \rowcolor{Gray} \net & ResNet101 & $512 \times 1408$ & \textbf{0.592} & \textbf{0.993} & \textbf{426} & \textbf{0.676} & \textbf{0.529} & \textbf{0.595} \\
        \midrule
        StreamPETR\textsuperscript{\textdaggerdbl}~\cite{wang2023exploring} & ViT-Adapter-B & $256 \times 704$ & 0.486 & 1.152 & 653 & 0.581 & 0.431 & 0.675 \\
        Sparse4Dv3\textsuperscript{\textdaggerdbl}~\cite{lin2023sparse4d} & ViT-Adapter-B & $256 \times 704$ & 0.499 & 1.075 & \textbf{445} & 0.592 & 0.448 & 0.615 \\
        \rowcolor{Gray} \net & ViT-Adapter-B & $256 \times 704$ & \textbf{0.555} & \textbf{1.030} & 502 & \textbf{0.601} & \textbf{0.492} & \textbf{0.587} \\
        \midrule
        SRCN3D\cite{shi2022srcn3d} & VoVNet-99 & $900 \times 1600$ & 0.439 & 1.280 & - & 0.545 & - & - \\ % from SRCN3D and Sparse4Dv3 paper
        ADA-Track~\cite{ding2024ada} & VoVNet-99 & $640 \times 1600$ & 0.479 & 1.246 & 767 & 0.602 & 0.430 & - \\
        QTrack\cite{yang2022quality} & VoVNet-99 & $640 \times 1600$ & 0.511 & 1.090 & 1144 & 0.585 & 0.465 & - \\ % from Sparse4Dv3 and QTrack paper
        ByteTrackV2~\cite{Zhang2023ByteTrackV22A} & VoVNet-99 & $640 \times 1600$ & 0.542 & 1.108 & 696 & - & 0.465 & - \\ % from ByteTrackV2 paper
        Sparse4Dv3\textsuperscript{\textdaggerdbl}~\cite{lin2023sparse4d} & VoVNet-99 & $640 \times 1600$ & 0.650 & 0.915 & 514 & 0.724 & 0.590 & 0.584 \\
        \rowcolor{Gray} \net & VoVNet-99 & $640 \times 1600$ & \textbf{0.663} & \textbf{0.885} & \textbf{426} & \textbf{0.732} & \textbf{0.604} & \textbf{0.547} \\
        \bottomrule
    \end{tabular}
    \footnotesize
    \textdaggerdbl:~Baselines trained with the code provided by the authors.
\end{threeparttable}
\end{table*}

\begin{table*}
\footnotesize
\centering
\caption{Results of 3D multi-object tracking on the nuScenes test set.}
\label{tab:results-track-test}
\setlength\tabcolsep{2.3pt}
\begin{threeparttable}
    \begin{tabular}{l | c c | c c c c c c }
        \toprule
        \textbf{Method} & \textbf{Backbone} & \textbf{Image Size} & \textbf{AMOTA$\uparrow$} & \textbf{AMOTP$\downarrow$} & \textbf{IDS$\downarrow$} & \textbf{Recall$\uparrow$} & \textbf{MOTA$\uparrow$} & \textbf{MOTP$\downarrow$} \\
        \midrule
        DORT~\cite{qing2023dort} & ConvNeXt-B & $640 \times 1600$ & 0.576 & 0.951 & 774 & 0.634 & 0.484 & 0.536 \\ % from nuScenes leaderboard
        RockTrack~\cite{li2024rocktrack} & InternImage-Base & - & 0.591 & 0.927 & 630 & 0.667 & 0.505 & 0.551 \\ % from nuScenes leaderboard
        SparseAD~\cite{zhang2024sparsead} & ViT-L & $800 \times 1600$ & 0.610 & 0.991 & 1144 & 0.706 & 0.549 & 0.594 \\ % from nuScenes leaderboard
        Sparse4Dv3~\cite{lin2023sparse4d} & ViT-L~\cite{fang2024eva} & $640 \times 1600$ & 0.643 & 0.820 & 699 & \textbf{0.743} & 0.593 & 0.481 \\
        StreamPETR~\cite{wang2023exploring} & ViT-L~\cite{fang2024eva} & $800 \times 1600$ & 0.653 & 0.876 & 1037 & 0.733 & 0.553 & 0.564 \\ % from nuScenes benchmark website and StreamPETR paper
        \rowcolor{Gray} \net & ViT-L~\cite{fang2024eva} & $640 \times 1600$ & \textbf{0.669} & \textbf{0.807} & \textbf{407} & 0.741 & \textbf{0.598} & \textbf{0.471} \\
        \bottomrule
    \end{tabular}
    \footnotesize
\end{threeparttable}
\end{table*}

%%%%%%%%%%%%%%%%%%%%%%%%%%%%%%%%%%%%%%%%%%%%%%%%%%%%%%%%%%%%%%%%%%%%%%%%%%%%%%%%

\subsection{Benchmarking Results}
We compare \net with state-of-the-art camera-based 3D detection and tracking methods across various backbones, resolutions, and datasets to demonstrate the generalization ability and scalability of our method. 

\subsubsection*{Detection results on nuScenes validation set}
In \cref{tab:results-3d-val}, we present the detection performance on the validation set. With a ResNet101 backbone at $512\times1408$ resolution, \net improves NDS by +0.2 and reduces mATE by -1.9 over Sparse4Dv3~\cite{lin2023sparse4d}. Using a ViT-Adapter-B~\cite{chen2023vision} backbone with a frozen DINOv2~\cite{oquab2023dinov2}, we observe a larger improvement of +2.2 NDS and +3.4 mAP, demonstrating that distillation is particularly effective when the backbone aligns with the distilled features. At higher resolution $640\times1600$ with a VoVNet-99~\cite{lee2019energy} backbone initialized with DD3D~\cite{park2021pseudo} weights, our method surpasses Sparse4Dv3 with a large improvement of +2.0 NDS and +2.4 mAP while reducing mATE by -3.2 and mASE by -1.4.

\subsubsection*{Detection results on nuScenes test set}
On the nuScenes test set, our model with a ViT-L EVA-02-initialized~\cite{fang2024eva} backbone slightly improves NDS over Sparse4Dv3 while substantially reducing mATE (-1.9), mASE (-0.8), and mAVE (-1.6), indicating more accurate localization and motion estimation, as shown in \cref{tab:results-3d-test}.

\subsubsection*{Tracking on nuScenes validation set}
From the results on the validation set presented in \cref{tab:results-track-val}, we observe that \net improves tracking quality consistently across backbones. With ResNet101 at $512\times1408$ resolution, the AMOTA score improves from 0.567 to 0.592 (+2.5), while AMOTP drops from 1.027 to 0.993 (lower is better), indicating tighter localization over matched tracks. Identity continuity also benefits, with IDS reduced from 557 to 426 (-131). With the ViT-Adapter-B at $256\times704$ resolution, we obtain a larger improvement in the AMOTA score from 0.499 to 0.555 (+5.6) compared to Sparse4Dv3 and +6.9 over StreamPETR. We also observe that the AMOTP score improves to 1.030 from 1.075. Although the IDS of 502 is slightly higher than Sparse4Dv3 which achieves 445, it is substantially lower than StreamPETR which achieves 653. 

With VoVNet-99 at $640\times1600$ resolution, the AMOTA score improves from 0.650 to 0.663 (+1.3) with a pronounced IDS reduction from 514 to 426 (-88) and lower AMOTP from 0.915 to 0.885. These trends indicate that (i) fusing PV and BEV strengthens temporal consistency with fewer ID switches and geometric precision with lower AMOTP, and (ii) the benefits persist from compact to high-capacity backbones, with the largest AMOTA improvement observed for ViT-Adapter-B with DINOv2 weights, where backbone features align with our DINOv2-guided BEV map.

\subsubsection*{Tracking on nuScenes test set}
The results on the test set presented in \cref{tab:results-track-test} show that \net achieves the best AMOTA score among camera-only methods with 0.669, surpassing StreamPETR by +1.6 and Sparse4Dv3 by +2.6. It also achieves the lowest AMOTP at 0.807 compared to 0.876 for StreamPETR and 0.820 for Sparse4Dv3, reflecting more precise trajectory localization. Most notably, IDS drops significantly to 407 from 699 compared to Sparse4Dv3 and 1037 of StreamPETR, evidencing substantially improved identity stability over long sequences. Together, these results confirm that our dual-view architecture with foundation model guided BEV maps enhances both detection quality along tracks (AMOTP) and the robustness of the association process (IDS), thereby establishing a new state of the art in vision-based joint 3D object detection and tracking.

\begin{table*}
\footnotesize
\centering
\caption{Long-range results of 3D object detection on the Argoverse 2 validation set.}
\label{tab:results-argoverse2-150m}
\setlength\tabcolsep{2.5pt}
\begin{threeparttable}
    \begin{tabular}{l | c c | c c | c c c c }
        \toprule
        \textbf{Method} & \textbf{Backbone} & \textbf{Image Size} & \textbf{CDS$\uparrow$} & \textbf{mAP$\uparrow$} & \textbf{mATE$\downarrow$} & \textbf{mASE$\downarrow$} & \textbf{mAOE$\downarrow$} \\
        \midrule
        % Results are copied from Far3D paper
        PETR~\cite{qing2023dort} & VoVNet-99 & $640 \times 960$ & 0.122 & 0.176 & 0.911 & 0.339 & 0.819 \\
        Sparse4Dv2~\cite{lin2023sparse4dv2} & VoVNet-99 & $640 \times 960$ & 0.134 & 0.189 & 0.832 & 0.343 & 0.723 \\
        StreamPETR~\cite{wang2023exploring} & VoVNet-99 & $640 \times 960$ & 0.146 & 0.203 & 0.843 & 0.321 & 0.650 \\
        RayDN~\cite{liu2024ray} & VoVNet-99 & $640 \times 960$ & 0.161 & 0.223 & 0.825 & 0.325 & 0.629 \\ % from RayDN paper
        Far3D~\cite{jiang2024far3d} & VoVNet-99 & $640 \times 960$ & 0.181 & 0.244 & 0.796 & 0.304 & \textbf{0.538} \\
        Sparse4Dv3\textsuperscript{\textdaggerdbl}~\cite{lin2023sparse4d} & VoVNet-99 & $640 \times 960$ & 0.186 & 0.247 & \textbf{0.692} & 0.302 & 0.558 \\
        \rowcolor{Gray} \net & VoVNet-99 & $640 \times 960$ & \textbf{0.198} & \textbf{0.264} & 0.734 & \textbf{0.299} & 0.573 \\
        \bottomrule
    \end{tabular}
    \footnotesize
    We evaluate across 26 object categories within a range of 150 meters. \textdaggerdbl:~Baselines trained by us with the code provided by the authors.
\end{threeparttable}
\end{table*}

\begin{table*}
\footnotesize
\centering
\caption{Near-range results of 3D object detection on the Argoverse 2 validation set.}
\label{tab:results-argoverse2-50m}
\setlength\tabcolsep{2.5pt}
\begin{threeparttable}
    \begin{tabular}{l | c c | c c | c c c c }
        \toprule
        \textbf{Method} & \textbf{Backbone} & \textbf{Image Size} & \textbf{CDS$\uparrow$} & \textbf{mAP$\uparrow$} & \textbf{mATE$\downarrow$} & \textbf{mASE$\downarrow$} & \textbf{mAOE$\downarrow$} \\
        \midrule
        Far3D~\cite{jiang2024far3d} & VoVNet-99 & $640 \times 960$ & 0.281 & 0.367 & 0.730 & 0.300 & \textbf{0.531} \\
        Sparse4Dv3\textsuperscript{\textdaggerdbl}~\cite{lin2023sparse4d} & VoVNet-99 & $640 \times 960$ & 0.294 & 0.381 & 0.706 & 0.314 & 0.666 \\
        \rowcolor{Gray} \net & VoVNet-99 & $640 \times 960$ & \textbf{0.313} & \textbf{0.406} & \textbf{0.642} & \textbf{0.287} & 0.565 \\
        \bottomrule
    \end{tabular}
    \footnotesize
    We evaluate across 26 object categories within a range of 50 meters. \textdaggerdbl:~Baselines trained by us with the code provided by the authors.
\end{threeparttable}
\end{table*}

\subsubsection*{Argoverse~2 long-range}
We further benchmark \net on Argoverse~2 at $\SI{150}{\meter}$ to assess long-range perception and cross-dataset generalization, as shown in \cref{tab:results-argoverse2-150m}. With a VoVNet-99 backbone, \net achieves higher CDS and mAP than Sparse4Dv3 (CDS +1.2, mAP +1.7). Notably, these results are obtained with a $\SI{50}{\meter}$ BEV grid range while relying on PV features beyond the BEV extent, demonstrating that our PV--BEV hybrid can maintain strong long-range performance. We restrict the BEV grid to $\SI{50}{\meter}$ range as a practical GPU memory tradeoff. BEV covers the near-range most critical for self-driving in city traffic, and PV features extend the perception range beyond the BEV extent without the memory overhead of larger BEV grids.

\subsubsection*{Argoverse~2 near-range}
We further evaluate models trained for $\SI{150}{\meter}$ long-range at a $\SI{50}{\meter}$ near-range setting. \net improves over Sparse4Dv3 on CDS (+1.9) and mAP (+2.5) and lowers mATE (-6.4) and mASE (-2.7), as summarized in \cref{tab:results-argoverse2-50m}. These results show that our method works especially well when foundation model guided BEV features are available. Moreover, combining BEV and PV features yields the benefits of both representations, rather than reverting to PV or BEV alone.

%%%%%%%%%%%%%%%%%%%%%%%%%%%%%%%%%%%%%%%%%%%%%%%%%%%%%%%%%%%%%%%%%%%%%%%%%%%%%%%%

\subsection{Ablation Study}

\begin{table*}
\footnotesize
\centering
\caption{Ablation study: BEV network architecture and DINOv2 map distillation.}
\label{tab:ablation-bev-map}
\setlength\tabcolsep{3.7pt}
\begin{threeparttable}
    \begin{tabular}{c c c c c | c c c}
        \toprule
        \textbf{PV DeformAgg.} & \textbf{BEV Net.} & \textbf{3D/BEV Data Aug.} & \textbf{Semantic Map} & \textbf{DINOv2 Map} & \textbf{NDS$\uparrow$} & \textbf{mAP$\uparrow$} & \textbf{AMOTA$\uparrow$} \\
        \midrule
        \checkmark & & &  & & 0.588 & 0.482 & 0.490 \\
        \checkmark & \checkmark & & & & 0.577 & 0.473 & 0.471 \\
        \checkmark & & \checkmark & & & 0.600 & 0.506 & 0.492 \\
        & \checkmark & \checkmark & & & 0.599 & 0.503 & 0.515 \\
        \checkmark & \checkmark & \checkmark & & & 0.615 & 0.520 & 0.515 \\
        \checkmark & \checkmark & \checkmark & \checkmark & & 0.611 & 0.519 & 0.534 \\
        \rowcolor{Gray}  \checkmark & \checkmark & \checkmark & & \checkmark & \textbf{0.619} & \textbf{0.527} & \textbf{0.555} \\
        \bottomrule
    \end{tabular}
    \footnotesize
    The baseline is Sparse4Dv3 with an image resolution of $256 \times 704$, ViT-Adapter-B as the backbone, and without instance denoising.
\end{threeparttable}
\end{table*}

\begin{table}
\footnotesize
\centering
\caption{DINOv2 BEV feature generation.}
\label{tab:ablation-dinov2-gen}
\setlength\tabcolsep{1.5pt}
\begin{threeparttable}
    \begin{tabular}{l | c c c}
        \toprule
        \textbf{Method} & \textbf{NDS$\uparrow$} & \textbf{mAP$\uparrow$} & \textbf{AMOTA$\uparrow$} \\
        \midrule
        Base - Static map & 0.611 & 0.520 & 0.522 \\
        \hspace{.5pt} + Dynamic objects & 0.609 & 0.521 & 0.548 \\
        \rowcolor{Gray} \hspace{.5pt} + Pointcloud accumulation & \textbf{0.619} & \textbf{0.527} & \textbf{0.555} \\
        %\rowcolor{Gray} \hspace{.5pt} + Multi-scale ensemble~\cite{voedisch2025pastel} & & & \\
        \bottomrule
    \end{tabular}
    \footnotesize
    % Image resolution of $256 \times 704$, ViT-Adapter-B as the backbone, without instance denoising.
\end{threeparttable}
\end{table}

We extensively evaluate the effect of various architectural and map supervision design choices. In \cref{tab:ablation-bev-map,tab:ablation-dinov2-gen,tab:ablation-dinov2-sup,tab:ablation-dinov2-loss,tab:ablation-oracle-lidar-dino}, we highlight the utilized variant in gray. We conducted all ablation studies on a small image resolution of $256\times704$, with a DINOv2 ViT-Adapter-B and without instance denoising from Sparse4Dv3 for efficiency reasons.

\subsubsection*{PV--BEV Network Architecture}

In \cref{tab:ablation-bev-map}, we show the effect of adding a BEV network together with BEV deformable aggregation in the head to the PV-only baseline. We observe that this addition reduces performance without 3D/BEV data augmentations. However, once we enable augmentations, both single-view settings become strong. With augmentations, PV-only achieves 0.600 NDS, 0.506 mAP, and 0.492 AMOTA, while BEV-only achieves 0.599 NDS, 0.503 mAP, and 0.515 AMOTA. We observe that detection is on par for both methods, while tracking is better for the BEV-only method (AMOTA +2.3). This indicates that, after we apply 3D/BEV augmentations, whether features come from PV or BEV is not crucial for detection, whereas BEV provides a unified metric space that benefits association in tracking. This is unexpected, as prior sparse query-based methods have traditionally relied exclusively on PV features. The results indicate that the strong performance of sparse query-based frameworks is primarily due to their effective query head design rather than the use of PV features themselves. When we combine PV and BEV, we observe the largest improvements of 0.615 NDS, 0.520 mAP, and 0.515 AMOTA. PV provides fine-grained, appearance-rich features, while BEV offers a spatially aligned metric space for multi-view fusion and motion reasoning. By jointly attending to both views, we leverage their complementary strengths.

\subsubsection*{DINOv2 Map Distillation}

In \cref{tab:ablation-bev-map}, we further study the effect of different supervision strategies for enriching BEV representations. Our ablations show that adding semantic BEV HD map supervision~\cite{liu2023bevfusion}, where rasterized semantic map classes are used to supervise the BEV feature map, provides an improvement in tracking performance (AMOTA +1.9) but does not improve detection. Additionally, this approach requires access to manually labeled HD maps during training. In contrast, our proposed supervision with DINOv2 BEV pseudo-labels yields improvements across both detection and tracking (NDS +0.4, mAP +0.7, AMOTA +4.0), while requiring only ground truth bounding boxes and object tracks, which are already available for the detection and tracking tasks. This demonstrates that the distillation of semantic and geometric features from foundation models provides a more scalable and effective way to enhance BEV representations than traditional HD map supervision. For fair comparison, we weight the semantic HD map supervision loss to match the scale of the DINOv2 distillation loss after 1000 training steps.

\begin{table}
\footnotesize
\centering
\caption{DINOv2 to BEV distillation supervision.}
\label{tab:ablation-dinov2-sup}
\setlength\tabcolsep{2.1pt}
\begin{threeparttable}
    \begin{tabular}{l | c c c}
        \toprule
        \textbf{Distillation to BEV Grid} & \textbf{NDS$\uparrow$} & \textbf{mAP$\uparrow$} & \textbf{AMOTA$\uparrow$} \\
        \midrule
        All cells & 0.608 & 0.512 & 0.500 \\
         \rowcolor{Gray} Cells with points & \textbf{0.619} & \textbf{0.527} & \textbf{0.555} \\
        \bottomrule
    \end{tabular}
    \footnotesize
    \end{threeparttable}
\end{table}

\begin{table}[t]
\footnotesize
\centering
\caption{DINOv2 to BEV distillation loss weighting.}
\label{tab:ablation-dinov2-loss}
\setlength\tabcolsep{2.5pt}
\begin{threeparttable}
    \begin{tabular}{l | c c c}
        \toprule
        \textbf{Loss Weight} & \textbf{NDS$\uparrow$} & \textbf{mAP$\uparrow$} & \textbf{AMOTA$\uparrow$} \\
        \midrule
        0.0 & 0.615 & 0.520 & 0.515 \\
        3.0 & 0.614 & 0.521 & 0.530 \\
        \rowcolor{Gray} 7.0 & \textbf{0.619} & \textbf{0.527} & \textbf{0.555} \\
        14.0 & 0.611 & 0.515 & 0.521 \\
        \bottomrule
    \end{tabular}
    \footnotesize
\end{threeparttable}
\footnotesize
\end{table}

\begin{table}[t]
\footnotesize
\centering
\caption{Oracle upper bound with LiDAR+DINOv2 point features.}
\label{tab:ablation-oracle-lidar-dino}
\setlength\tabcolsep{2.5pt}
\begin{threeparttable}
    \begin{tabular}{l | c | c c c}
        \toprule
        \textbf{Method} & \textbf{GT Tracks} & \textbf{NDS$\uparrow$} & \textbf{mAP$\uparrow$} & \textbf{AMOTA$\uparrow$} \\
        \midrule
        \multicolumn{5}{l}{\textit{\net (oracle setting)}} \\
        [.25ex]
        LiDAR + DINOv2 & \cmark & \textbf{0.772} & \textbf{0.761} & \textbf{0.705} \\
        LiDAR only     & \cmark & 0.731 & 0.690 & 0.699 \\
        \midrule
        \multicolumn{5}{l}{\textit{SOTA camera-LiDAR baselines}} \\
        [.25ex]
        MV2DFusion~\cite{wang2025mv2dfusion} & \xmark & 0.754 & 0.739 & — \\
        DINO-MOT~\cite{lee2024dino} & \xmark & — & — & 0.803 \\
        \bottomrule
    \end{tabular}
    \footnotesize
    Oracle rows replace the camera BEV network with a LiDAR BEV encoder and use sequence point cloud accumulation that requires GT tracks; external baselines do not.
\end{threeparttable}
\end{table}

\subsubsection*{DINOv2 BEV Feature Generation}
In \cref{tab:ablation-dinov2-gen}, we compare the different parts of the scene point cloud, which is used to generate the DINOv2 BEV pseudo-labels. Generating the pseudo-label point cloud only for a single frame and excluding dynamic objects using the GT boxes does not improve over Sparse4Dv3. Our method improves detection and tracking only when we include dynamic objects and accumulate the point cloud of the static scene, while separately accumulating each dynamic object in its own object-centric coordinates.

\subsubsection*{DINOv2 to BEV Distillation Supervision}
In \cref{tab:ablation-dinov2-sup}, we investigate the effect of supervising all BEV grid cells with the distillation loss or only cells with LiDAR points. As shown, it is crucial to only supervise cells that contain points to avoid learning any LiDAR point cloud pattern in the camera BEV features.

\subsubsection*{DINOv2 to BEV Distillation Loss Weighting}
Finally, we provide results for different distillation loss weights \(\lambda_{\text{distill}}\) in \cref{tab:ablation-dinov2-loss}, showing that it is crucial to set it appropriately. Setting this weight too high can result in negative transfer for detection and tracking due to the multi-task setting.

\begin{table*}
\footnotesize
\centering
\caption{Results in varying weather and illumination conditions.}
\label{tab:analysis-weather}
\setlength\tabcolsep{2.5pt}
\begin{threeparttable}
    \begin{tabular}{l | c c c | c c c | c c c }
        \toprule
        & \multicolumn{3}{c|}{\textbf{Day}} & \multicolumn{3}{c|}{\textbf{Rain}} & \multicolumn{3}{c}{\textbf{Night}} \\
        \textbf{Method} & \textbf{NDS$\uparrow$} & \textbf{mAP$\uparrow$} & \textbf{AMOTA$\uparrow$} & \textbf{NDS$\uparrow$} & \textbf{mAP$\uparrow$} & \textbf{AMOTA$\uparrow$} & \textbf{NDS$\uparrow$} & \textbf{mAP$\uparrow$} & \textbf{AMOTA$\uparrow$} \\
        \midrule
        Sparse4Dv3 & 0.594 & 0.487 & 0.512 & 0.642 & 0.551 & 0.535 & 0.351 & \textbf{0.319} & \textbf{0.511} \\
        \rowcolor{Gray} \net & \textbf{0.610} & \textbf{0.515 } & \textbf{0.548} & \textbf{0.666} & \textbf{0.580} & \textbf{0.603} & \textbf{0.375} & 0.306 & 0.483 \\
        \bottomrule
    \end{tabular}
    \footnotesize
    The baseline is Sparse4Dv3 with an image resolution of $256 \times 704$, ViT-Adapter-B as the backbone, and without instance denoising.
\end{threeparttable}
\end{table*}

\begin{figure*}
    \centering

    % First visualization
    \subfloat{
        \includegraphics[width=\textwidth]{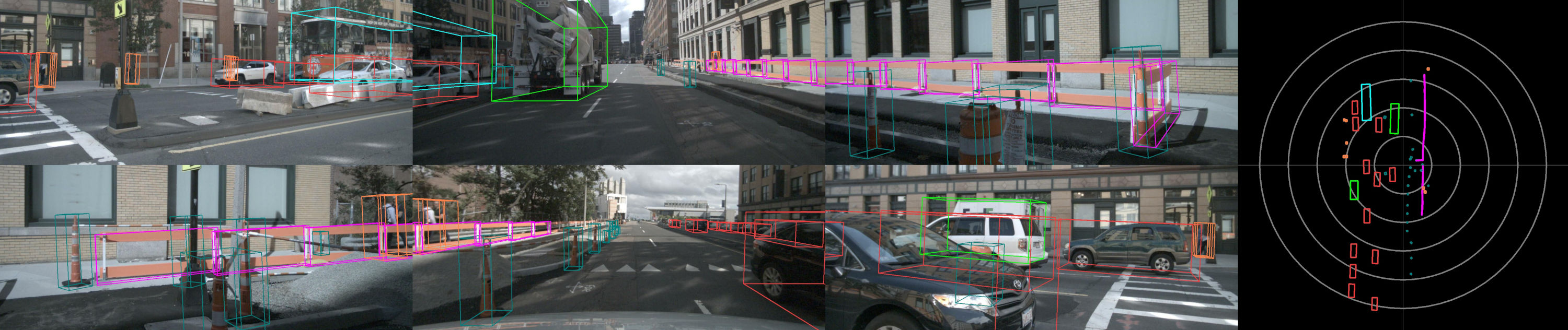}
        %\caption{}
    }
    \vspace{-.2cm}

    % Second visualization
    \subfloat{
        \includegraphics[width=\textwidth]{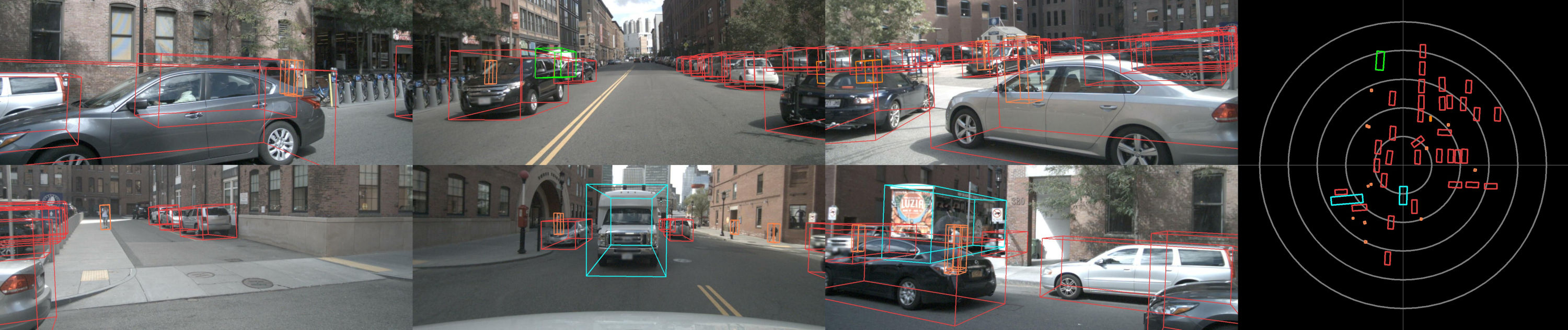}
        %\caption{}
    }
    \vspace{-.2cm}

    % Third visualization
    \subfloat{
        \includegraphics[width=\textwidth]{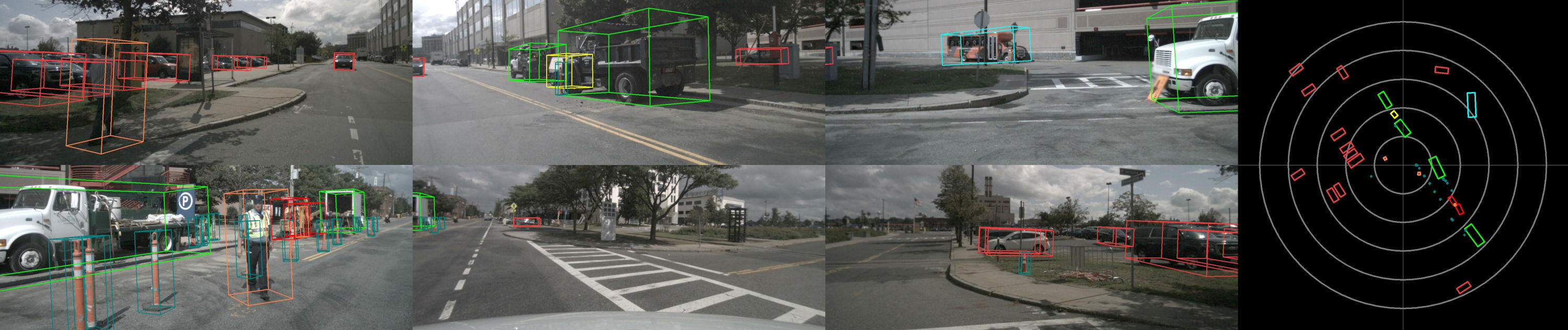}
        %\caption{}
    }
    \vspace{-.2cm}

    \caption{
    Visualization of 3D object detection results from our proposed \net with ViT-L backbone on the nuScenes test set. Classes are color-coded as follows: \textcolor[rgb]{0.93,0.23,0.23}{\rule{0.25cm}{0.25cm}} car,
\textcolor[rgb]{0,1,0}{\rule{0.25cm}{0.25cm}} truck,
\textcolor[rgb]{1,0,0}{\rule{0.25cm}{0.25cm}} construction vehicle,
\textcolor[rgb]{0,1,1}{\rule{0.25cm}{0.25cm}} bus,
\textcolor[rgb]{1,1,0}{\rule{0.25cm}{0.25cm}} trailer,
\textcolor[rgb]{1,0,1}{\rule{0.25cm}{0.25cm}} barrier,
\textcolor[rgb]{1,1,1}{\rule{0.25cm}{0.25cm}} motorcycle,
\textcolor[rgb]{1,0.5,0}{\rule{0.25cm}{0.25cm}} bicycle,
\textcolor[rgb]{1,0.51,0.28}{\rule{0.25cm}{0.25cm}} pedestrian,
\textcolor[rgb]{0,0.5,0.5}{\rule{0.25cm}{0.25cm}} traffic cone.
    }
    \label{fig:qualitative-nuscenes-test}
\end{figure*}

\subsubsection*{Oracle upper bound}
\label{ssec:result-abl-oracle}
To motivate and demonstrate the potential of our distillation approach, we evaluate an oracle setting where the sequence-accumulated point cloud used to form pseudo-labels is provided as input to the network, either as raw 3D points or augmented with a DINOv2 feature vector per point. Note that sequence accumulation uses ground truth tracks, which are not available during inference. Points are voxelized, per-voxel features are extracted with an MLP, the resulting 3D voxel tensor is then converted into a stack of pillars by using max-pooling along the z-axis, and a lightweight 2D CNN encodes the pillar features into BEV~\cite{yuan2024presight}. These BEV features are fed into the head via BEV deformable aggregation while the camera-based BEV network is disabled. 

As shown in \cref{tab:ablation-oracle-lidar-dino}, our method with a DINOv2-augmented LiDAR point cloud as input achieves 0.772 NDS, 0.761 mAP, and 0.705 AMOTA, while using only the LiDAR point cloud without additional per-point features achieves 0.731 NDS, 0.690 mAP, and 0.699 AMOTA. Notably, the \emph{LiDAR + DINOv2} oracle surpasses the state-of-the-art LiDAR-camera detector MV2DFusion~\cite{wang2025mv2dfusion}, while a gap remains to the best tracking method DINO-MOT~\cite{lee2024dino}. These results indicate that LiDAR provides precise 3D geometry, whereas DINOv2 contributes descriptive features that benefit both detection and tracking. This strong oracle further motivates our camera-only DINOv2 to BEV distillation, as transferring foundation-model features into BEV recovers part of the oracle gains without requiring LiDAR or ground truth tracks at inference.

%%%%%%%%%%%%%%%%%%%%%%%%%%%%%%%%%%%%%%%%%%%%%%%%%%%%%%%%%%%%%%%%%%%%%%%%%%%%%%%%

\subsection{Weather and Illumination Robustness Analysis}

We evaluate the robustness of our approach under different environmental conditions in \cref{tab:analysis-weather} by using the nuScenes splits introduced by~\cite{schramm2024bevcar}. \net consistently outperforms Sparse4Dv3 across day, night, and rain scenarios. The largest improvements are observed in rainy conditions (NDS +2.4, AMOTA +6.8), highlighting the benefit of foundation model guided BEV maps in providing semantic cues when the image quality deteriorates. At night, the improvements are smaller, but \net still achieves a higher NDS score, demonstrating that enriched BEV representations contribute to more reliable detection under challenging illumination.

%%%%%%%%%%%%%%%%%%%%%%%%%%%%%%%%%%%%%%%%%%%%%%%%%%%%%%%%%%%%%%%%%%%%%%%%%%%%%%%%

\subsection{Qualitative Results}

\begin{figure*}
    \centering

    % First visualization
    \subfloat{
        \includegraphics[width=0.8\textwidth]{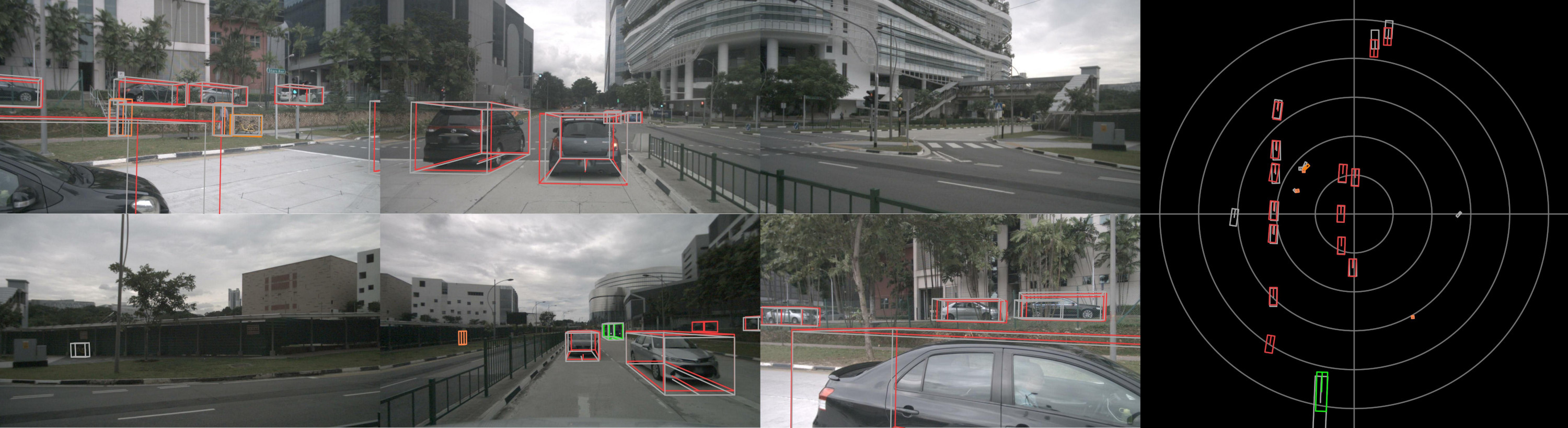}
    }
    \vspace{-.2cm}

    % Second visualization
    \subfloat{
        \includegraphics[width=0.8\textwidth]{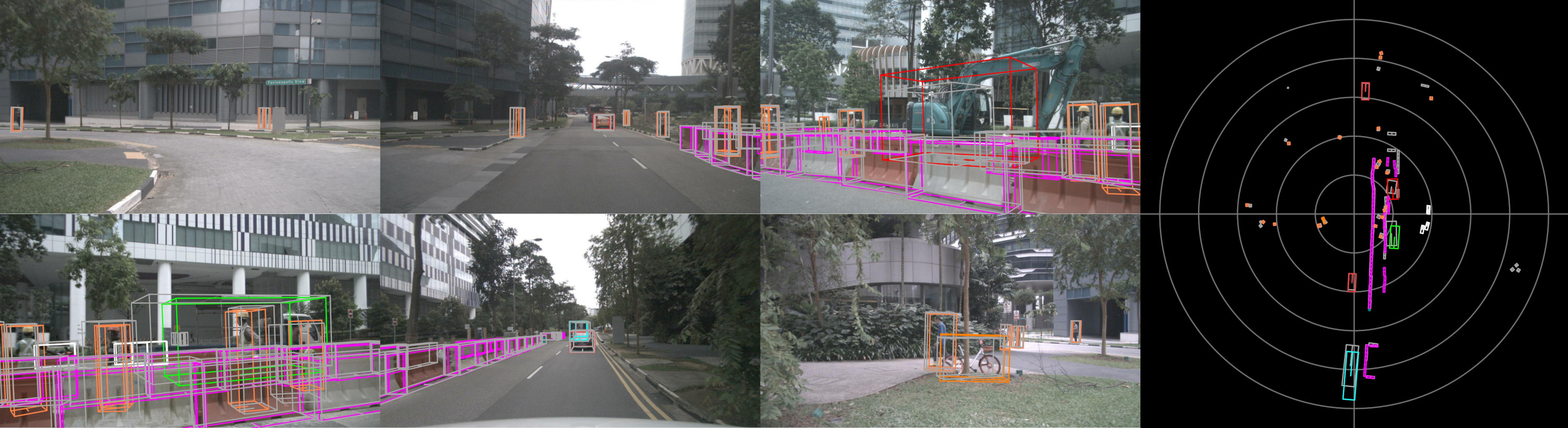}
    }
    \vspace{-.2cm}

    % Third visualization
    \subfloat{
        \includegraphics[width=0.8\textwidth]{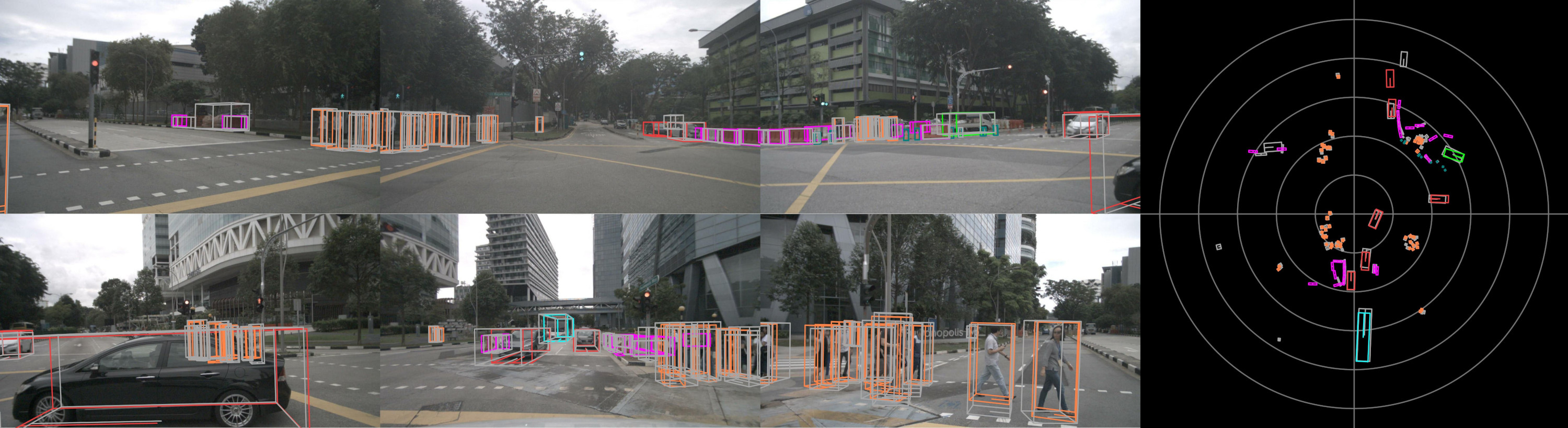}
    }
    \vspace{-.2cm}

    \caption{
    Visualization of 3D object detection results from our proposed \net with VoVNet-99 backbone on the nuScenes validation set. Ground truths are colored in gray while classes are color-coded as follows: \textcolor[rgb]{0.93,0.23,0.23}{\rule{0.25cm}{0.25cm}} car,
\textcolor[rgb]{0,1,0}{\rule{0.25cm}{0.25cm}} truck,
\textcolor[rgb]{1,0,0}{\rule{0.25cm}{0.25cm}} construction vehicle,
\textcolor[rgb]{0,1,1}{\rule{0.25cm}{0.25cm}} bus,
\textcolor[rgb]{1,1,0}{\rule{0.25cm}{0.25cm}} trailer,
\textcolor[rgb]{1,0,1}{\rule{0.25cm}{0.25cm}} barrier,
\textcolor[rgb]{1,1,1}{\rule{0.25cm}{0.25cm}} motorcycle,
\textcolor[rgb]{1,0.5,0}{\rule{0.25cm}{0.25cm}} bicycle,
\textcolor[rgb]{1,0.51,0.28}{\rule{0.25cm}{0.25cm}} pedestrian,
\textcolor[rgb]{0,0.5,0.5}{\rule{0.25cm}{0.25cm}} traffic cone.
    }
    \label{fig:qualitative-nuscenes-validation}
\end{figure*}

\begin{figure*}
    \centering

    % First visualization
    \subfloat{
        \includegraphics[width=1.0\textwidth]{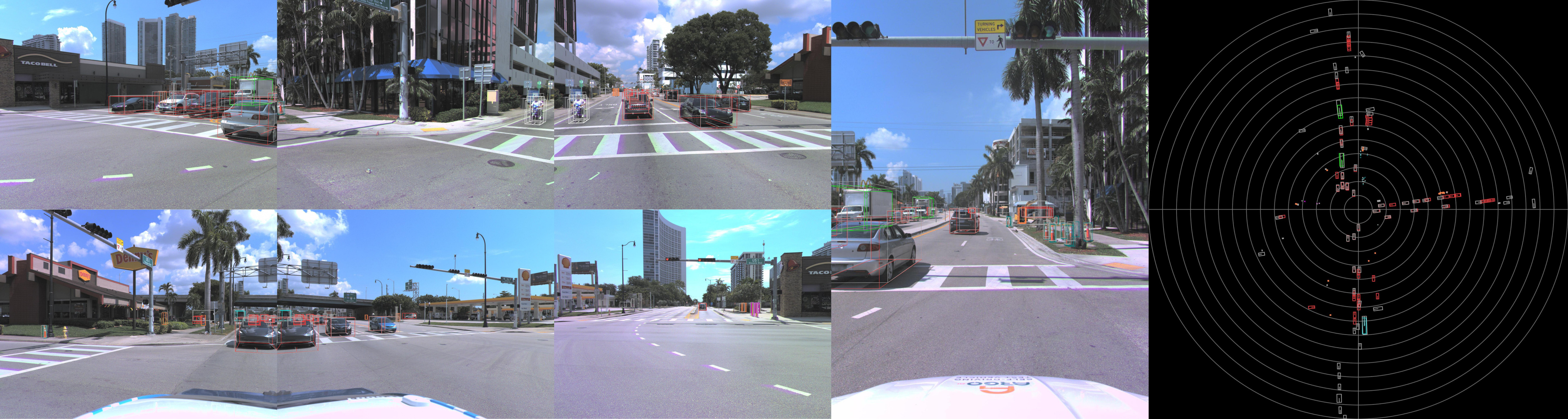}
        %\caption{}
    }
    \vspace{-.2cm}

    % Second visualization
    \subfloat{
        \includegraphics[width=1.0\textwidth]{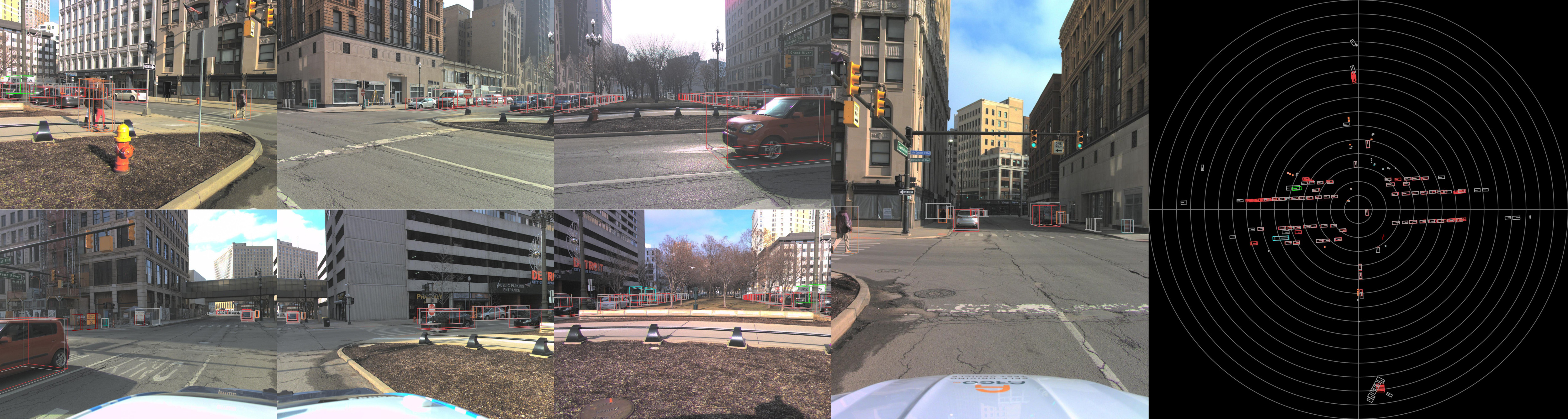}
        %\caption{}
    }
    \vspace{-.2cm}

    \caption{
    Visualization of long-range ($\SI{150}{\meter}$) 3D object detection results from our proposed \net with VoVNet-99 backbone on the Argoverse 2 validation set. Ground truths are colored in gray while classes are color-coded as follows:\;
    \textcolor[rgb]{0.93,0.23,0.23}{\rule{0.25cm}{0.25cm}} car\; %(REGULAR\_VEHICLE),\;
    \textcolor[rgb]{0,1,0}{\rule{0.25cm}{0.25cm}} truck,\; %(TRUCK/BOX\_TRUCK/TRUCK\_CAB),\;
    \textcolor[rgb]{1,0,0}{\rule{0.25cm}{0.25cm}} large vehicle,\; %(LARGE\_VEHICLE),\;
    \textcolor[rgb]{0,1,1}{\rule{0.25cm}{0.25cm}} bus,\; %(BUS/ARTICULATED\_BUS/SCHOOL\_BUS),\;
    \textcolor[rgb]{1,1,0}{\rule{0.25cm}{0.25cm}} trailer,\; %(VEHICULAR\_TRAILER/MESSAGE\_BOARD\_TRAILER),\;
    \textcolor[rgb]{1,0,1}{\rule{0.25cm}{0.25cm}} barrier,\; %(BOLLARD),\;
    \textcolor[rgb]{1,1,1}{\rule{0.25cm}{0.25cm}} motorcycle,\; %(MOTORCYCLE/MOTORCYCLIST),\;
    \textcolor[rgb]{1,0.50,0}{\rule{0.25cm}{0.25cm}} bicycle,\; %(BICYCLE/BICYCLIST),\;
    \textcolor[rgb]{1,0.51,0.28}{\rule{0.25cm}{0.25cm}} pedestrian,\;
    \textcolor[rgb]{0,0.50,0.50}{\rule{0.25cm}{0.25cm}} construction cone/barrel,\;
    \textcolor[rgb]{0,0,0}{\rule{0.25cm}{0.25cm}} dog,\;
    \textcolor[rgb]{0,0,1}{\rule{0.25cm}{0.25cm}} mobile pedestrian crossing sign,\;
    \textcolor[rgb]{0.50,0,0}{\rule{0.25cm}{0.25cm}} sign,\;
    \textcolor[rgb]{0,0.50,0}{\rule{0.25cm}{0.25cm}} stop sign,\;
    \textcolor[rgb]{0,0,0.50}{\rule{0.25cm}{0.25cm}} stroller,\;
    \textcolor[rgb]{0.50,0.50,0.50}{\rule{0.25cm}{0.25cm}} wheelchair,\;
    \textcolor[rgb]{0.50,0.50,0}{\rule{0.25cm}{0.25cm}} wheeled device,\;
    \textcolor[rgb]{0.50,0,0.50}{\rule{0.25cm}{0.25cm}} wheeled rider.
    }
    \label{fig:qualitative-argoverse2-validation}
\end{figure*}

\begin{figure*}
    \centering

    % First visualization
    \subfloat[Sparse4Dv3 (VoVNet-99)]{
        \includegraphics[width=\textwidth]{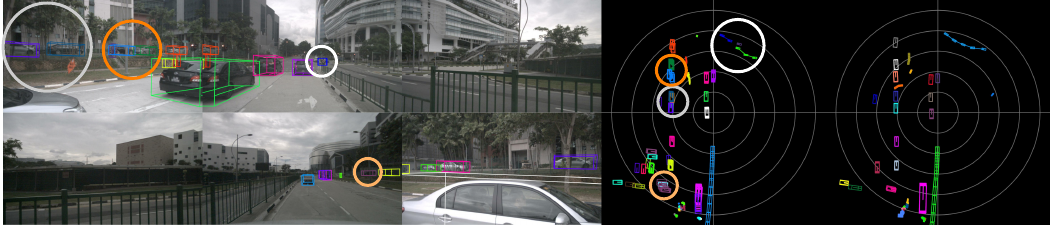}
    }
    \vspace{-.2cm}

    % Second visualization
    \subfloat[\net (VoVNet-99)]{
        \includegraphics[width=\textwidth]{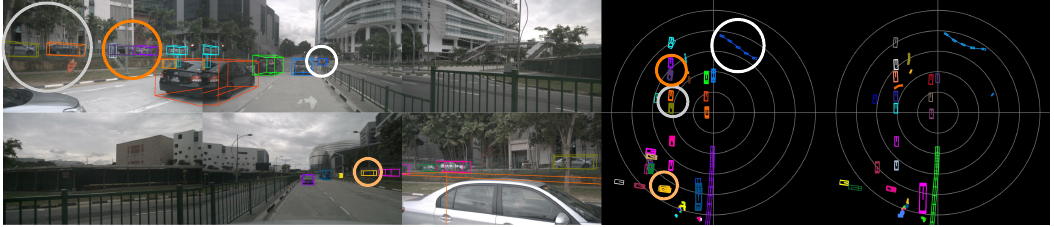}
    }

    \caption{
    Visualization of 3D multi-object tracking results from Sparse4Dv3 and our proposed \net with VoVNet-99 backbone on the nuScenes validation set. Our method shows substantially fewer ID switches (white/gray) and more accurate localization of 3D bounding boxes (orange). The BEV visualizations on the right show the ground truth tracks.
    }
    \label{fig:qualitative-nuscenes-validation-tracking}
\end{figure*}

In \cref{fig:qualitative-nuscenes-test}, we present qualitative results of 3D object detection from our \net (ViT-L) model on challenging scenes from the nuScenes test set. We visualize detections projected onto both the perspective view (PV) and bird’s-eye view (BEV) to highlight the model’s ability to accurately localize and classify objects across diverse scenes, including crowded urban environments and scenes with complex occlusions. Powered by a ViT-L backbone, our largest variant provides the strongest performance with accurate, stable detections and improved box orientations, as can be seen in the barrier headings in the first visualization. The foundation model guided BEV maps further reduce the average translation error of objects, thus improving object localization. Our model also detects rare truck categories with high reliability, including a concrete mixer in the first visualization and several trucks in the last visualization. This indicates that large backbones pre-trained on large 2D detection datasets improve the detection of long-tail classes. Finally, our method remains robust in scenes with many objects and strong occlusions, as the second visualization shows a densely populated parking lot where detections are stable despite heavy occlusions.

In \cref{fig:qualitative-nuscenes-validation}, we show additional detections from \net with a VoVNet-99 backbone on the nuScenes validation set. Predictions are colored by class while ground truths are overlaid in gray. We observe that the model reliably detects objects that are not extremely occluded. Vehicles are localized accurately at both near and far ranges, while pedestrians are typically localized precisely at near-range but can show increased depth uncertainty at long-range. The model remains stable in scenes with many diverse, highly dynamic traffic participants.

In \cref{fig:qualitative-argoverse2-validation}, we show the detections from \net with a VoVNet-99 backbone on the Argoverse~2 validation set. At $\SI{50}{\meter}$ near-range, the model performs similarly to nuScenes. Objects are localized accurately and stably across diverse scenes. The model also handles partial occlusions well, maintaining correct class labels and boxes when objects are partly hidden by nearby traffic participants or the static environment. Beyond the near-range, the detections remain reliable up to roughly $\SI{100}{\meter}$, though we observe depth uncertainty between $\SI{60}{\meter}$ and $\SI{100}{\meter}$ that can result in duplicated detections along the viewing ray or small distance deviations. At distances greater than $\SI{100}{\meter}$, the performance degrades as objects are very small in the images and often heavily occluded. These conditions are also challenging for human annotators. Using higher-resolution images may mitigate these far-range failures by providing more details for long-range objects.

In \cref{fig:qualitative-nuscenes-validation-tracking}, we compare 3D multi-object tracking between Sparse4Dv3 and \net, both with VoVNet-99 backbone. On the left, 3D bounding boxes are colored by track ID. On the right, the BEV visualization shows the predicted and ground truth tracks. \net yields fewer ID switches and lower inter-frame jitter, particularly for long-range objects. For example, Sparse4Dv3 produces two ID switches for the motorcycle in front of the ego vehicle and some for the parked cars on the left, while our method maintains consistent identities. Moreover, Sparse4Dv3 shows larger frame-to-frame centroid drift for parked cars at long ranges, whereas \net keeps these boxes temporally stable.
These qualitative results complement the quantitative findings and underscore the benefit of our hybrid design. By combining PV and BEV features and distilling semantic-rich BEV maps from foundation models, \net achieves reliable detection performance across diverse scenes and object densities. 

\section{Limitations}
\label{sec:limitatios}
While \net achieves strong performance in both detection and tracking, it has several limitations. First, our method relies on LiDAR data during training to project DINOv2 features into 3D space for DINOv2 BEV pseudo-label generation. Although LiDAR is not required at inference time, this dependence may limit applicability in purely camera-only pipelines. Future work could explore alternatives such as monocular or multi-view depth estimation or 3D reconstruction methods~\cite{wang2025moge, wang2025vggt} to generate point clouds for supervision during training. Second, our current distillation strategy pools high-dimensional DINOv2 features into BEV space, which may lead to information loss. Retaining features in full 3D before projection could offer richer representations, but this would come at a significant memory cost due to the high-dimensional features.

\section{Conclusion}
\label{sec:conclusion}
We presented \net, a state-of-the-art architecture for joint 3D object detection and multi-object tracking. \net's novel approach to distill features from vision foundation models into BEV to guide the learning of BEV map representations boosts object detection and tracking performance on the nuScenes and Argoverse 2 benchmarks. Furthermore, \net's design of downstream task heads benefits both from fine-grained image features and unified BEV feature maps in a hybrid manner. Our empirical study shows that rich BEV  representations are a crucial modality for autonomous driving systems, and when utilized properly, it can significantly enhance the performance of key tasks in autonomous driving pipelines, namely 3D object detection and multi-object tracking. Our Argoverse~2 results at both near- and long-range further confirm these findings, highlighting generalization across datasets and long-range perception. Addressing the identified limitations offers promising directions for extending \net towards broader and more reliable perception in autonomous driving systems.

%%%%%%%%%%%%%%%%%%%%%%%%%%%%%%%%%%%%%%%%%%%%%%%%%%%%%%%%%%%%%%%%%%%%%%%%%%%%%%%%

%\FloatBarrier
\footnotesize
\bibliographystyle{IEEEtran}
\bibliography{references}

%%%%%%%%%%%%%%%%%%%%%%%%%%%%%%%%%%%%%%%%%%%%%%%%%%%%%%%%%%%%%%%%%%%%%%%%%%%%%%%%

%\newpage
%\input{sections/7_supplementary}

%%%%%%%%%%%%%%%%%%%%%%%%%%%%%%%%%%%%%%%%%%%%%%%%%%%%%%%%%%%%%%%%%%%%%%%%%%%%%%%%

\end{document}